\documentclass[journal]{IEEEtran}
\ifCLASSINFOpdf
\else
\fi
\usepackage{times}
\usepackage{epsfig}
\usepackage{graphicx}
\usepackage{amsmath}
\usepackage{amssymb}
\usepackage{hyperref}
\usepackage{subfigure}
\usepackage{amsthm}

\usepackage{multirow}
\usepackage{booktabs}
\usepackage{algorithm}
\usepackage{algorithmic}
\usepackage{epstopdf}
\usepackage{color}

\begin{document}
%
\title{Instance-Aware Hashing for Multi-Label Image Retrieval}
%
%
%

\author{Hanjiang~Lai,
        Pan~Yan,
        Xiangbo Shu,Yunchao~Wei, Shuicheng~Yan, ~\IEEEmembership{Senior Member,~IEEE}
\thanks{Hanjiang Lai is with School of Data and Computer Science, Sun Yat-Sen University, China,  e-mail: (laihanj@gmail.com).}
\thanks{Yan Pan is with School of Data and Computer Science, Sun Yat-Sen University, Guangzhou, 510006, e-mail: (panyan5@mail.sysu.edu.cn). Yan Pan is Corresponding author.}
\thanks{Xiangbo Shu is with School of Computer Science and Technology, Nanjing, China. e-mail: (shuxb104@gmail.com).}
\thanks{Yunchao Wei is with the Institute of Information Science, Beijing Jiaotong University, e-mail: (wychao1987@gmail.com).}
\thanks{Shuicheng Yan is with Department
of Electrical and Computer Engineering, National University of Singapore,  e-mail: (eleyans@nus.edu.sg).}
}

%
%

\markboth{IEEE Transactions on Image Processing}%
{Shell \MakeLowercase{\textit{et al.}}: Bare Demo of IEEEtran.cls for Journals}
%



\maketitle

\begin{abstract}
Similarity-preserving hashing is a commonly used method for nearest neighbour search in large-scale image retrieval. For image retrieval, deep-networks-based hashing methods are appealing since they can simultaneously learn effective image representations and compact hash codes. This paper focuses on deep-networks-based hashing for multi-label images, each of which may contain objects of multiple categories. In most existing hashing methods, each image is represented by one piece of hash code, which is referred to as semantic hashing. This setting may be suboptimal for multi-label image retrieval. To solve this problem, we propose a deep architecture that learns \textbf{instance-aware} image representations for multi-label image data, which are organized in multiple groups, with each group containing the features for one category. The instance-aware representations not only bring advantages to semantic hashing, but also can be used in  category-aware hashing, in which an image is represented by multiple pieces of hash codes and each piece of code corresponds to a category. Extensive evaluations conducted on several benchmark datasets demonstrate that, for both semantic hashing and category-aware hashing, the proposed method shows substantial improvement over the state-of-the-art supervised and unsupervised hashing methods.
\end{abstract}

\begin{IEEEkeywords}
Multi-Label, Image Retrieval, Instance-Aware Image Representation, Category-Aware Hashing, Semantic Hashing, Deep Learning.
\end{IEEEkeywords}

%
\IEEEpeerreviewmaketitle

\section{Introduction}
%
%
%
%
\IEEEPARstart{L}{arge}-scale image retrieval, which is to find images containing similar objects as in a query image, has attracted increasing interest due to the ever-growing amount of available image data on the Web. Similarity-preserving hashing is a popular nearest neighbor search technique for image retrieval on datasets with millions or even billions of images.

\begin{figure}
  \centering
  \includegraphics[width=0.95\hsize]{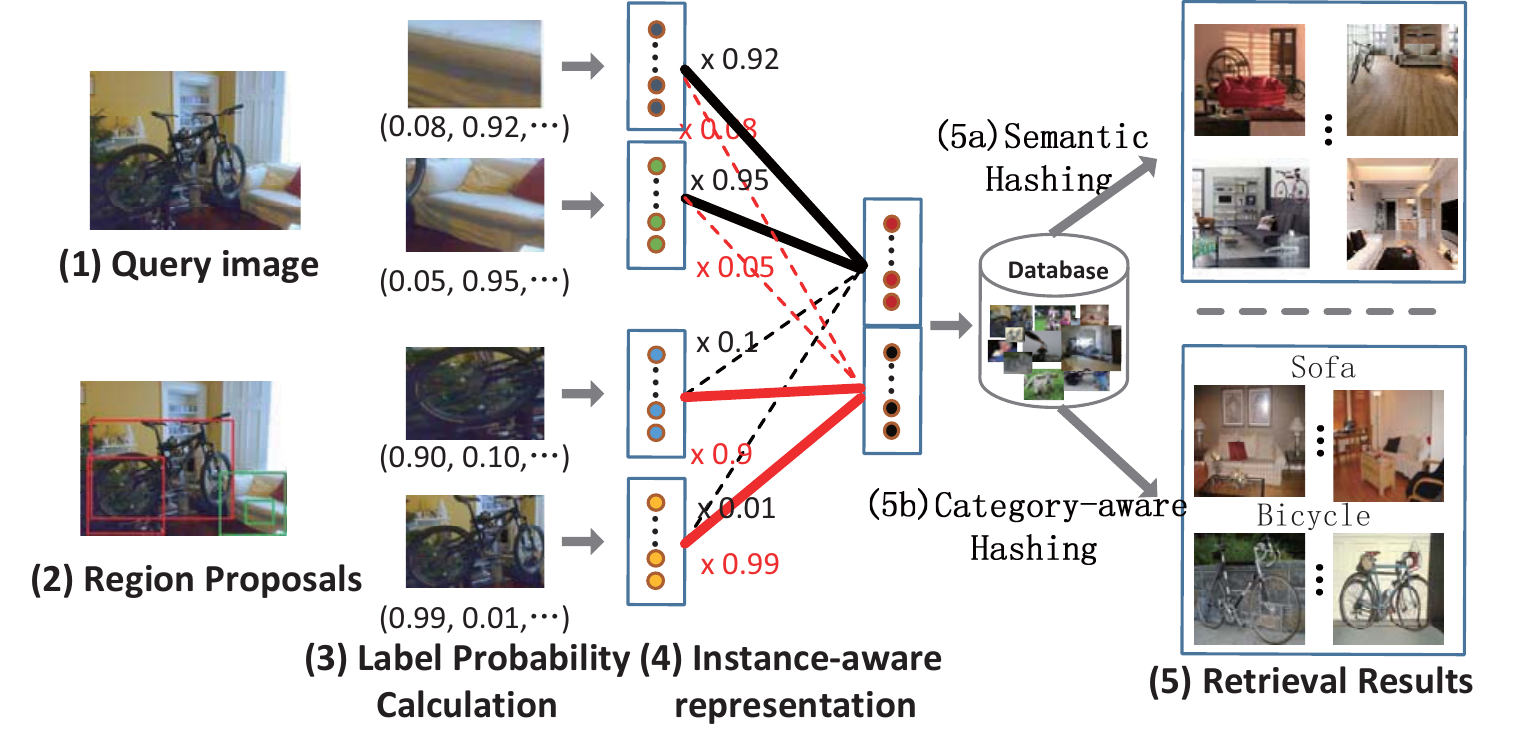}
  \caption{Illustration of instance-aware image retrieval. (1) Given a query image, e.g., containing  a bicycle and a sofa, the proposal method (2) generates region proposals,  (3) computes the label probability scores for each proposal, (4) encodes each proposal to an intermediate feature vector, and then computes the weighted average of these vectors (with the label probability scores being the weights) to generate the instance-aware representations organized in multiple groups, each corresponding to an object. After that, this representation is converted to (5a) one piece of hash code for semantic hashing or (5b) multiple pieces of hash codes, each piece corresponding to a category, for category-aware hashing.
 }
 \label{example}
\end{figure}

A representative stream of similarity-preserving hashing is learning-to-hash, i.e., learning to compress data points (e.g., images) into binary representations such that semantically similar data points have nearby binary codes. The existing learning-to-hash methods can be divided into two main categories: unsupervised methods and supervised methods. Unsupervised methods (e.g.,~\cite{ITQ,KLSH,AGH}) learn a set of hash functions from unlabeled data without any side information. Supervised methods (e.g.,~\cite{CNNH,KSH,BRE,TSH}) try to learn compact hash codes by leveraging supervised information on data points (e.g., similarities on pairs of images).
Among various supervised learning-to-hash methods for image retrieval, an emerging stream is deep-networks-based hashing that learns bitwise codes as well as image representations via carefully designed deep neural networks. Several deep-networks-based hashing methods have been proposed (e.g.,~\cite{CNNH,onestep,zhao2015deep}).

\begin{figure*}
  \centering
  \includegraphics[width=0.9\hsize]{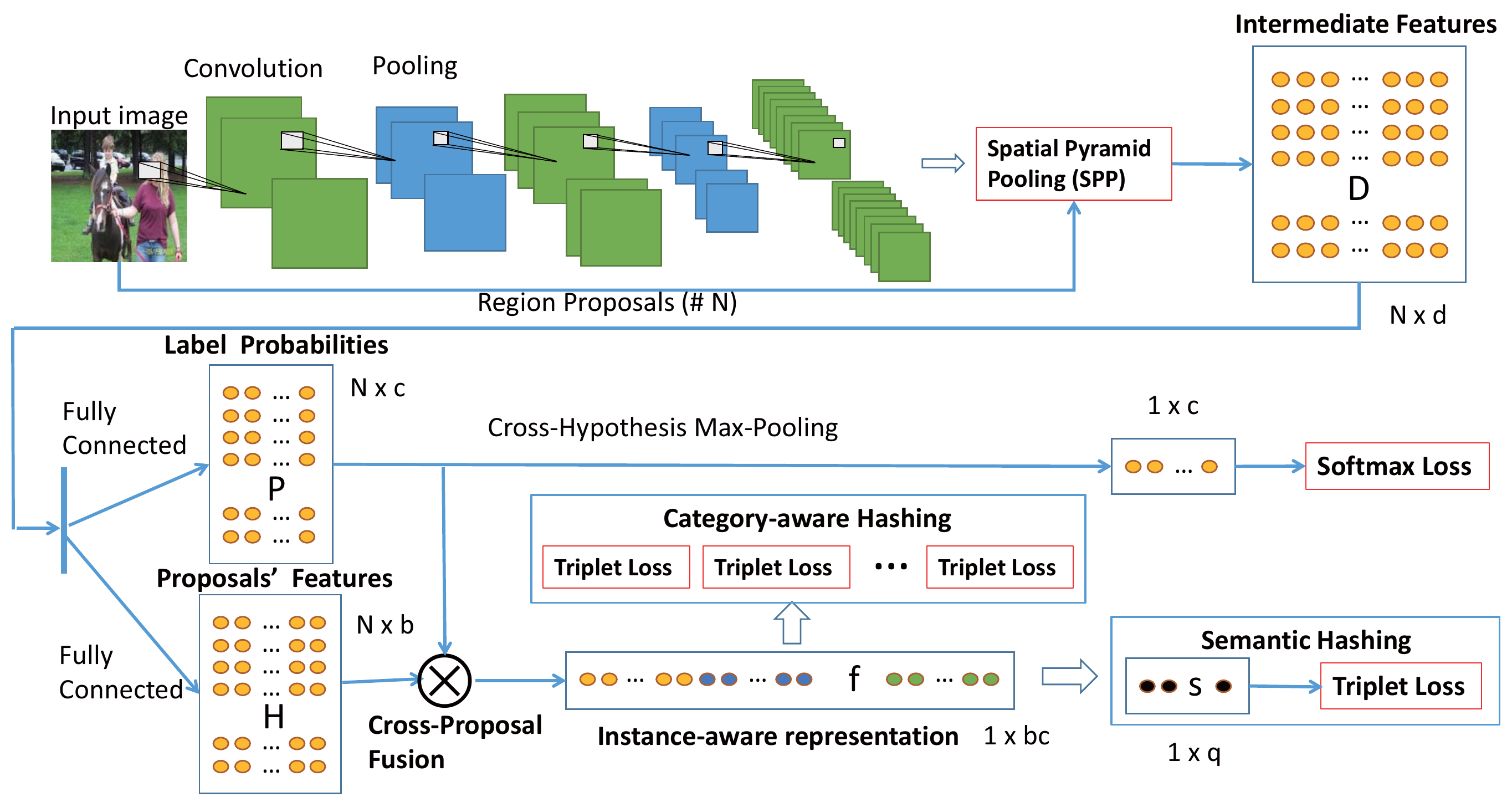}
  \caption{Overview of the proposed deep architecture for hashing on multi-label images. The proposed architecture takes an image (e.g, in $c$ classes) and its automatically generated $N$ region proposals as inputs. The image firstly goes trough the deep convolution sub-network, and then $N$ intermediate feature vectors are generated for the region proposals via the spatial pyramid pooling scheme. With the intermediate features, our network is divided into two branches: one for calculating the label probabilities of the region proposals (see Fig.~3), and the other for generating the proposals' features. Cross-proposal fusion is performed to merge the proposals' features (with label probabilities) into an intermediate multiple-slice representation (i.e., $f$) in which each slice corresponds to one category (see Fig.~4). After that, this intermediate representation is converted to multiple pieces of hash codes (for category-aware hashing) or one piece of hash code (for semantic hashing).}
    \label{overview}
\end{figure*}

Multi-label images, each of which may contain objects of multiple categories, are widely involved in many image retrieval systems. However, in most existing hashing methods for images, the semantic similarities are defined at image level, and each image is represented by one piece of hash code. This setting may be suboptimal for multi-label image retrieval.

In this paper, we consider \textbf{instance-aware retrieval} for multi-label image data, which includes semantic hashing~\cite{semantichashing} and category-aware hashing. Specifically, given a multi-label query image, 
a natural demand is to organize the retrieved results in groups, each group corresponding to one category. For example, as shown in Figure~\ref{example}, given a query image containing a \textit{bicycle} and a \textit{sofa}, one would like to organize the retrieved results in two groups: each image in the first (second) group contains a bicycle (sofa) similar to the one in the query image.
In order to achieve instance-aware retrieval, we propose a new image representation organized in groups, by incorporating automatically generated candidate object proposals and label probability calculation into the learning process. Figure~\ref{example} shows an example for the generation of the instance-aware image representation. 

More specifically, we propose a deep neural network that simultaneously learns binary hash codes and the representation tailored for multi-label image data. As shown in Figure~\ref{overview}, the proposed architecture has four building blocks: 1) a set of $N$ automatically generated candidate object proposals in the form of bounding boxes, as inputs to the deep neural network; 2) stacked convolutional layers to capture the features of the input proposals, followed by a Spatial Pyramid Pooling (SPP) layer~\cite{SPP-Net} to map each proposal to a $d$-dimensional intermediate representation; 3) a label probability calculation
 module that maps the intermediate representation to the image labels (in $c$ classes), which leads to an $N\times c$ probability matrix with the $i$-th row representing the label probabilities of the $i$-th proposal belonging to each class; 4) a hash coding module, where firstly an instance-aware representation is captured, in which the probability matrix in the third module is used as the input, and then either category-aware hash codes or semantic hash codes are generated based on this representation.

The proposed deep architecture can be used to generate hash codes for category-aware hashing, where an image is represented by multiple pieces of hash codes, each of which corresponds to a category. In addition, we show that the proposed image representation can improve the quality of semantic hashing in which an image is represented by one piece of hash code.


Our contributions in this paper can be summarized as follows. First, we propose a deep architecture that can generate hash codes for instance-aware retrieval. To the best of our knowledge, we are the first to conduct instance-aware retrieval via learning-based hashing. Second, we propose to incorporate automatically generated candidate object proposals and label probability calculation in the proposed deep architecture. We empirically show that the proposed method has superior performance gains over several state-of-the-art hashing methods.

\section{Related Work}
Due to the encouraging search speed, hashing has
become a popular method for nearest neighbor search in large-scale
image retrieval.

Hashing methods can be divided into
data independent hashing and data dependent hashing. The early
efforts mainly focus on data independent hashing. For example, the
notable Locality-Sensitive Hashing (LSH)~\cite{LSH} method
constructs hash functions by random projections or random
permutations that are independent of the data points. The main
limitation of data independent methods is that they usually
require long hash codes to obtain good performance. However, long
hash codes lead to inefficient search due to the required large storage space and the
low recall rates.

Learning-based hashing (or Learning-to-hash) pursues a compact
binary representation from the training data.
Based on whether side information is used or not, learning-to-hash
methods can be divided into two categories: unsupervised methods
and supervised methods.

Unsupervised methods try to learn a set of similarity-preserving
hash functions only from the unlabeled data. Representative methods in this category include Kernelized LSH
(KLSH)~\cite{KLSH}, Semantic
hashing~\cite{semantic}, Spectral
hashing~\cite{SH}, Anchor Graph Hashing~\cite{AGH}, and Iterative Quantization (ITQ)~\cite{ITQ}.
 Kernelized LSH
(KLSH)~\cite{KLSH} generalizes LSH to accommodate arbitrary kernel
functions, making it possible to learn hash functions which preserve
data points' similarity in a kernel space. Semantic
hashing~\cite{semantic} generates hash functions by a deep
auto-encoder via stacking multiple restricted Boltzmann machines
(RBMs). Graph-based hashing methods, such as Spectral
hashing~\cite{SH} and Anchor Graph Hashing~\cite{AGH}, learn
non-linear mappings as hash functions which try to preserve the
similarities within the data neighborhood graph. In order to reduce
the quantization errors, Iterative Quantization (ITQ)~\cite{ITQ}
seeks to learn an orthogonal rotation matrix which is applied to the
data matrix after principal component analysis projections.

Supervised methods aim to learn better bitwise representations by
incorporating supervised information. Notable methods in this category include
Binary Reconstruction
Embedding (BRE)~\cite{BRE}, Minimal Loss Hashing (MLH)~\cite{MLH}, Supervised Hashing with Kernels (KSH)~\cite{KSH}, Column Generation
Hash (CGHash)~\cite{CGHash}, and
Semi-Supervised Hashing (SSH)~\cite{SSH}.
Binary Reconstruction
Embedding (BRE)~\cite{BRE} learns hash functions by explicitly
minimizing the reconstruction errors between the original distances
of data points and the Hamming distances of the corresponding binary
codes. Minimal Loss Hashing (MLH)~\cite{MLH} learns
similarity-preserving hash codes by minimizing a hinge-like loss
function which is formulated as structured prediction with latent
variables. Supervised Hashing with Kernels (KSH)~\cite{KSH} is a
kernel-based supervised method which learns to hash the data points
to compact binary codes whose Hamming distances are minimized on
similar pairs and maximized on dissimilar pairs. Column Generation
Hash (CGHash)~\cite{CGHash} is a column generation based method to
learn hash functions with proximity comparison information.
Semi-Supervised Hashing (SSH)~\cite{SSH} learns hash functions via
minimizing similarity errors on the labeled data while
simultaneously maximizing the entropy of the learnt hash codes over
the unlabeled data. In most image retrieval applications, the number of labeled positive samples is small, which results in bias towards the negative samples and over-fitting. Tao \textit{et al.}~\cite{tao2006asymmetric} proposed an asymmetric bagging and random subspace SVM (ABRS-SVM) to handle these problems.

In supervised hashing methods for image retrieval, an emerging
stream is the deep-networks-based
methods~\cite{torralba2008small,CNNH,onestep,zhao2015deep} which learn image
representations as well as binary hash codes. Xia \textit{et al.}~\cite{CNNH} proposed Convolutional-Neural-Networks-based Hashing (CNNH), which is a two-stage method. In its first stage, approximate hash codes are learned from the supervised information. Then, in the second stage, hash functions are learned based on those approximate hash codes via deep convolutional networks. Lai \textit{et al.}~\cite{onestep} proposed a one-stage hashing method that generates bitwise hash codes via a carefully designed deep architecture. Zhao \textit{et al.}~\cite{zhao2015deep} proposed a ranking based hashing method for learning hash
functions that preserve multi-level semantic similarity between images, via deep convolutional networks. 
Lin \textit{et al.}~\cite{lin2015deep} proposed to learn the hash codes and image representations in a point-wised manner, which is suitable for large-scale datasets. Wang \textit{et al.}~\cite{wangdeep} proposed Deep Multimodal Hashing with Orthogonal Regularization (DMHOR) method for multimodal data.
All of these methods generate one piece of hash code for each image, which may be inappropriate for multi-label image retrieval. Different from the existing methods, the proposed method can generate multiple pieces of hash codes for an image, each piece corresponding to a(n) instance/category.

\section{The Proposed Method}



Our method consists of four modules. The first module is to generate region proposals for an input image. The second module is to capture the features for the generated region proposals. It contains a deep convolution sub-network followed by a Spatial Pyramid Pooling layer~\cite{SPP-Net}. The third module is a
label probability calculation 
  module, which outputs a probability matrix whose $i$-th row represents the probability scores of the $i$-th proposal belonging to each class. The fourth module is a hash coding module that firstly generates the instance-aware representation, and then converts this representation to hash codes for either category-aware hashing or semantic hashing. In the following, we will present the details of these modules, respectively.

\subsection{Region Proposal Generation Module\label{RPG}}

 Many methods for generating category-independent region proposals have been proposed, e.g., Constrained Parametric Min-Cuts (CPMC)~\cite{carreira2012cpmc}, Selective Search~\cite{uijlings2013selective}, Multi-scale Combinatorial Grouping (MCG)~\cite{mcg}, BInarized Normed Gradients (BING)~\cite{bing} and Geodesic Object Proposals (GOP)~\cite{GOP}. In this paper, we use GOP~\cite{GOP} to automatically generate region proposals for an input. Note that other methods for region proposal generation can also be used in our framework.
 
GOP is a method that can generate both segmentation masks and bounding box proposals. We use the code\footnote{http://www.philkr.net/home/gop} provided by the authors to generate the bounding boxes for region proposals.

\subsection{Deep Convolution Sub-Network Module}
GoogLeNet~\cite{GoogleLeNet} is
a recently proposed deep architecture that has shown its success in
object categorization and object detection. The core of GoogLeNet is
the Inception-style convolution module which allows increasing the depth and width of the network while
keeping reasonable computational costs. Here we adopt the
architecture of GoogLeNet as our basic framework to compute the
features for the input proposals. Since the GoogLeNet is a very deep network and has many layers, we use the pre-trained GoogLeNet model\footnote{http://dl.caffe.berkeleyvision.org/bvlc\_googlenet.caffemodel} to initialize its weights which can be regarded as regularization~\cite{difficulty} and help  its generalization.

However, since the number of
generated region proposals for an input image may be large (e.g.,
more than 1000), it is computationally expensive if one directly
uses GoogLeNet to extract features from these proposals. This is
unaffordable for hashing-based retrieval since the retrieval system
may need a long time to respond to a query.

To address this issue, we use the ``Spatial Pyramid Pooling" (SPP)
scheme~\cite{SPP-Net}. The advantage of using SPP is that we can compute
the feature map from the entire input image only once. Then, with
this feature map, we pool features in each generated region proposal
to generate a fixed-length representation. Using SPP, we avoid
repeatedly computing features for the input region proposals via a
deep convolutional network. Specifically, as shown in Figure~\ref{overview}, we add an SPP layer after the last convolutional
layer of GoogLeNet. We assume that each input image has $N$
automatically generated region proposals. For each input region
proposal, we encode its top-left and bottom-right coordinates to a
$4$-dimensional vector $\bold{L}_i \in \mathbb{R}^4 (i=1,2,...,N)$.
The elements in this vector are scaled to $[0,1]$ by dividing the
width/height of the image, making them invariant to the absolute
image size. With this $4$-dimensional vector as the input, the SPP layer
generates a fixed-length feature vector for the corresponding
proposal. Through the SPP layer, we assume that each proposal is
mapped to a $d$-dimensional intermediate feature vector. Hence, for
each input image, the output of this module is an $N \times d$
matrix.

After this module, the network is divided into two branches: one for
the label probability calculation module, and the other for the hash
coding module.

\begin{figure}
  \centering
  \includegraphics[width=0.9\hsize]{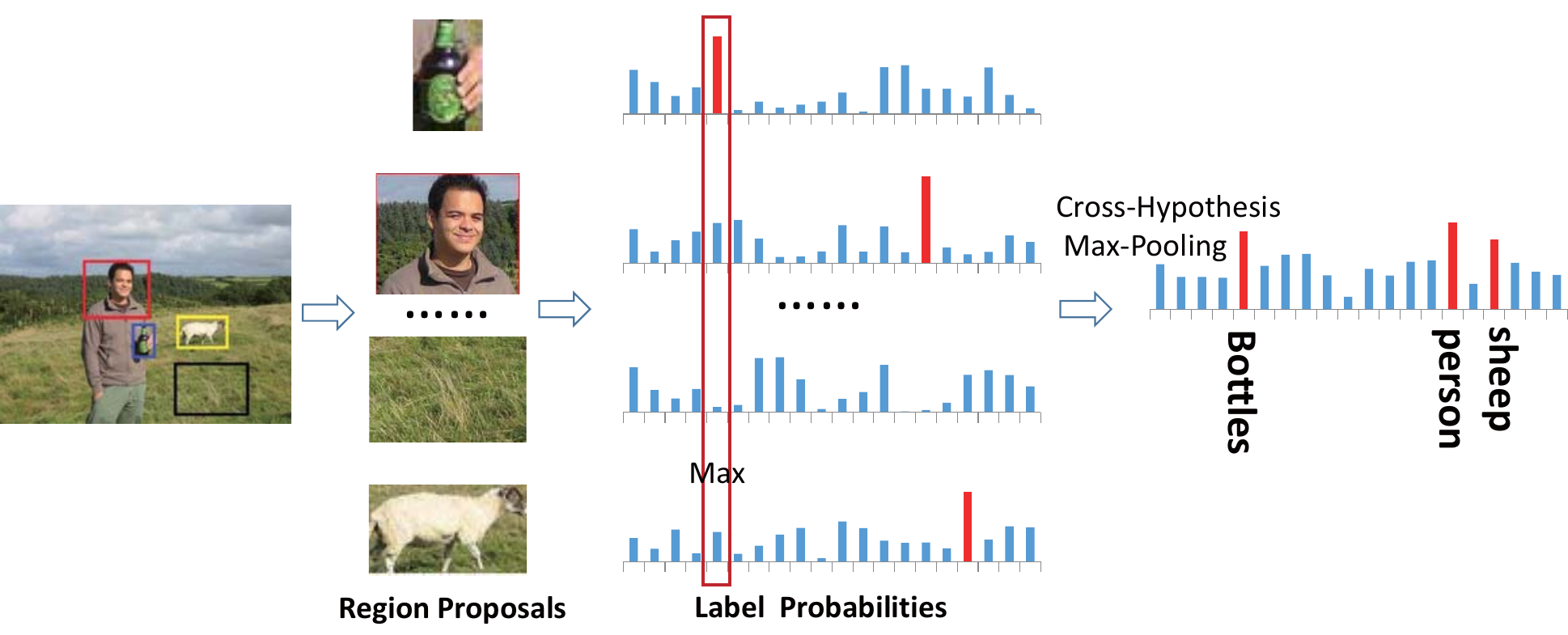}
  \caption{Illustration of the proposed label probability calculation module. For an image (in $c$ classes) with $N$ region proposals, our network generates a probability vector for each proposal, e.g., $M^i \in \mathbb{R}^c$ ($i=1,\cdots,N$). After that, the cross-hypothesis max-pooling is used to fuse the $N$ probability vectors into one vector.} 
    \label{lpcm}
\end{figure}

\subsection{Label Probability Calculation Module\label{MLCM}}

In this subsection, we will show how to learn the label probability for each region proposal. Suppose there are $c$ class labels, and a probability vector is generated for each proposal, e.g., $P^i = (P_1^i,\cdots,P_c^i)$ indicates that the probability of the image containing the $j$-th category is $P_j^i$.  

However, we do not have the ground truth labels for each proposal. Thus probability distribution can not be directly learned. Fortunately, in the multi-label image annotation, there is a label for the whole image, e.g., $(I,Y)$, where $I$ represents an image and $Y$ is the ground truth label. $Y \in \mathbb{R}^c$ and $Y_{j} \in \{1,0\}, j = 1,\cdots,c$. $Y_{j}$ is equal to 1 if the $j$-th label is relevant to image $I$ and 0 for the irrelevant case. Hence, we can firstly fuse the $N$ proposals into one and then use the whole image's label to learn as shown in Figure~\ref{lpcm}.

More specifically, with the $N\times d$ matrix $\bold{D}$ in which the $i$-th row
$\bold{D}^i$ represents the $d$-dimensional intermediate feature
for the $i$-th proposal, in this module, we first use a
fully-connected layer to compress $\bold{D}^i$ to a $c$-dimensional
vector $\bold{M}^i\in \mathbb{R}^c$ $(i=1,2,...,N)$.

After that, we use the cross-hypothesis
max-pooling~\cite{wei2014cnn} to fuse
$\bold{M}^1,\bold{M}^2,...,\bold{M}^N$ to one $c$-dimensional
vector. Specifically, let $\bold{M}$ be the $N$ by $c$ matrix whose
$i$-th row is $\bold{M}^i$. The cross-hypothesis max-pooling can be
formulated as
\begin{equation}
\bold{m}_j = \max \{ \bold{M}^1_j, \bold{M}^2_j, \cdots,
\bold{M}^N_j \}, \forall j = 1, \cdots, c,
\end{equation}
where $\bold{m}_j$ is the pooled value that corresponds to the
$j$-th category. 

Using $\bold{m}_j$ $(j=1,2,...,c)$, we calculate a probability
distribution $\bold{p}=(\bold{p}_1,\bold{p}_2,...,\bold{p}_c)$ expressed by
\begin{equation}
   \bold{p}_{j} = \frac{ \exp (\bold{m}_j)}{\sum_{k=1}^c \exp
   (\bold{m}_k)},
\end{equation}
where $\bold{p}_j$ can be regarded as the probability score that the
input image contains an object in the $j$-th category. Using such cross-hypothesis max-pooling, if the $i$-th proposal contains the $j$-th category, then the output $\bold{p}_{j}$ should have a large value and $\bold{M}^i_j$ will have a high response. Hence, it can guide the learning of $\bold{M}$.

In this module, we define a loss function based on the cross entropy
between the probability scores and the ground truth labels:

\begin{equation}
    \ell_C=- \sum_{j \in c_{+}} \frac{1}{|c_{+}|} \log (\bold{p}_{j}),
\end{equation}
where we denote $c_{+}$ as the set of categories which the input
image belongs to, and $|c_{+}|$ as the number of elements in
$c_{+}$. This loss function is also referred to as Softmax-Loss~\cite{gong2013deep}, which is a widely used loss function in the Convolutional neural networks. The (sub-)gradients with respect to $\bold{m}_j$ are
\begin{equation}
\frac{\partial \ell_C}{\partial \bold{m}_j} = \left\{
\begin{array}{rl}
&\bold{p}_j - \frac{1}{|c_{+}|}, \ \quad \quad if \ Y_j = 1 \\
&\bold{p}_j, \ \quad \quad \quad \quad \quad  if \ Y_j = 0  \\
\end{array}.
\right.
\end{equation}
It can be easily integrated in back propagation in neural networks. 

%

After that, similarly to $\bold{p}$, we can define a probability matrix $\bold{P}$ for the region proposals
 as
 \begin{equation}
 \bold{P}^i_{j} =  \frac{ \exp (\bold{M}^i_j)}{\sum_{k=1}^c \exp (\bold{M}^i_k)}(
i=1,\cdots,N),
\end{equation}
 where $\bold{P}^i_{j}$ represents the $j$-th element in the $i$-th row of $\bold{P}$. $\bold{P}^i_{j}$ can be viewed as the probability that
the $i$-th proposal contains an object of the $j$-th category.

\subsection{Hash Coding Module}
In this subsection, we will show how to convert the image representation into (a) one piece of hash code for semantic hashing or (b) multiple pieces of hash codes, each piece corresponding to a category, for category-aware hashing.

With the $N\times d$ matrix $\bold{D}$ as the input, in this module, we first use a
fully-connected layer to compress each $\bold{D}^i$ to a $b$-dimensional
vector $\bold{H}^i\in \mathbb{R}^b$ $(i=1,2,...,N)$, where $\bold{H}^i$ corresponds to the $i$-th proposal. We denote $\bold{H}$ as the $N$ by $b$ matrix whose $i$-th row is $\bold{H}^i$.

\subsubsection{Cross-Proposal Fusion}
In order to convert $\bold{H}$ into the instance-aware representation of the input image, we propose a \textit{cross-proposal fusion} strategy, by using the probability matrix $\bold{P}$ from the label probability calculation module.

Specifically, with the $N$ by $b$ feature matrix $\bold{H}$ and the $N$ by $c$ matrix $\bold{P}$ where $\bold{P}^i_{j}$ represents the probability of the $i$-th region proposal belonging to the $j$-th category, we fuse $\bold{H}$ and $\bold{P}$ into a long vector with $c\times b$ elements. This vector is organized in $c$ groups, each group representing $b$-dimensional features corresponding to one category.

Let $\bold{H}^i$, $\bold{P}^i$ represent the $i$-th row of $\bold{H}$, $\bold{P}$, respectively. The cross-proposal fusion can be formulated as
\begin{equation}
\bold{f}=\frac{1}{N}\sum_{i=1}^{N}\bold{P}^i\otimes \bold{H}^i,
\end{equation}
where $\otimes$ is the Kronecker product. For the $c$-dimensional vector $\bold{P}^i$ and the $b$-dimensional vector $\bold{H}^i$, the Kronecker product $\bold{P}^i \otimes \bold{H}^i$ is a $(c\times b)$-dimensional vector:
\begin{displaymath}
\begin{split}
 (&\bold{P}^i_1\bold{H}^i_1,\bold{P}^i_1\bold{H}^i_2,...,\bold{P}^i_1\bold{H}^i_b,\\&\bold{P}^i_2\bold{H}^i_1,\bold{P}^i_2\bold{H}^i_2,...,\bold{P}^i_2\bold{H}^i_b,\\&...\\&\bold{P}^i_c\bold{H}^i_1,\bold{P}^i_c\bold{H}^i_2,...,\bold{P}^i_c\bold{H}^i_b). \end{split}
 \end{displaymath}

Let $\bold{f}=(\bold{f}^{(1)},\bold{f}^{(2)},...,\bold{f}^{(c)})$, where $\bold{f}^{(j)}$ is a $b$-dimensional vector. It is easy to verify that
\begin{displaymath}
\bold{f}^{(j)}=\frac{1}{N}\sum_{i=1}^{N}\bold{P}^i_j \bold{H}^i.
\end{displaymath}
Since $\bold{H}^i$ represents the features of the $i$-th proposal, $\bold{f}^{(j)}$ can be regarded as the weighted average of the proposals' features. If the $i$-th proposal has a relatively higher/lower score $\bold{P}^i_j$ (meaning that the $i$-th proposal likely/unlikely belongs to the $j$-th category), the feature vector $\bold{H}^i$ (associated to the $i$-th proposal) has more/less contribution to the weighted average $\bold{f}^{(j)}$.




\begin{figure}
  \centering
  \includegraphics[width=0.9\hsize]{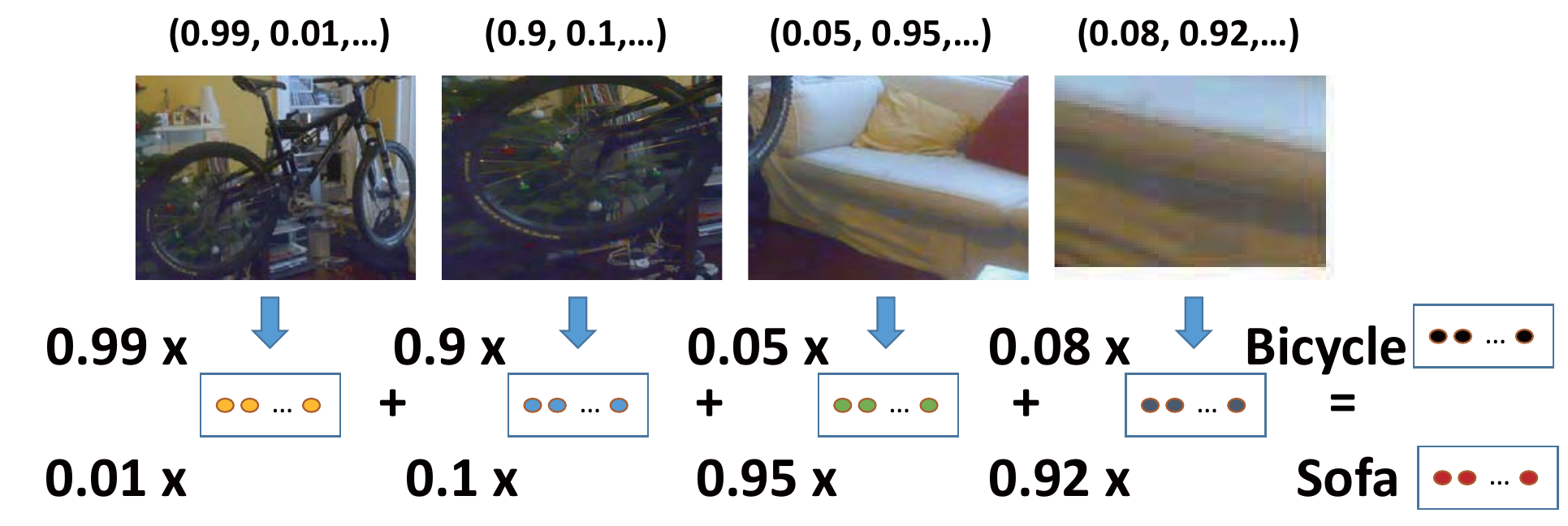}
  \caption{Illustration of the cross-proposal fusion. Each region proposal is firstly encoded into a feature vector. And these feature vectors are fused into an intermediate feature representation, by using the probability scores (learned from the label probability calculation module) as the weights.}
    \label{cross_proposal_fusion}
\end{figure}

Figure~\ref{cross_proposal_fusion} shows an illustrative example of cross-proposal fusion. Suppose there are only $2$ categories (\textit{bicycle }and \textit{sofa}, $c=2$ ) in all of the images. For the input image, $4$ region proposals are generated in the first module ($N = 4$). Then, suppose  the label probability calculation module generates a $4\times 2$ probability matrix $\bold{P}$ with $\bold{P}^1 = (0.99, 0.01)$, $\bold{P}^2 =
(0.9, 0.1)$, $\bold{P}^3 = (0.05, 0.95)$ and $\bold{P}^4 = (0.08, 0.92)$. For the $1$st proposal, $\bold{P}^1 = (0.99, 0.01)$ indicates that it is very likely to contain a bicycle (with a score $0.99$), but it seems unlikely to contain a sofa (with a score $0.01$). In the hash coding module, the $i$-th proposal is represented by a $b$-dimensional feature vector $\bold{H}^i$. Finally, for the input image, we conduct cross-proposal fusion to obtain an instance-aware representation $\bold{f}=(\bold{f}^{(1)}, \bold{f}^{(2)})$, where the representation of ``\textit{bicycle}'' is $\bold{f}^{(1)}=\frac{1}{4}(0.99\bold{H}^{1}  + 0.9\bold{H}^{2} + 0.05\bold{H}^{3} + 0.08\bold{H}^{4})$, and the representation of ``\textit{sofa}'' is  $\bold{f}^{(2)}=\frac{1}{4}(0.01\bold{H}^{1}  + 0.1\bold{H}^{2} + 0.95\bold{H}^{3} + 0.92\bold{H}^{4})$.

The cross-proposal fusion is a crucial step for the instance-aware image representation. If the image contains an object, then the instance-aware representation will have a high response for the object. It also gives us a simple way to combine the multi-label information into the hashing procedure. 

\textbf{Discussions.} A concern arising here is that some input proposals may be inaccurate or even do not contain any objects, which will make the features generated by these proposals noisy and harm the final performance. We argue that, to some extent, the operations in the proposed Cross-Hypothesis Max-Pooling and the Cross-Proposal Fusion can reduce the negative effects of the possibly noisy input proposals. Firstly, in the label probability calculation module before the cross-hypothesis max-pooling, each input proposal is assigned with a set of probability scores (one score for one label). Higher scores are assigned to the proposals that may contain objects with more confidence, and lower scores are assigned to those noisy proposals. Hence, those noisy proposals may more probably be suppressed by the cross-hypothesis max-pooling. Secondly, similar to~\cite{liu2014classification}, in the cross-proposal fusion, the proposal' features are weighted by their probability scores. Hence, those noisy proposals' features have less contribution to the final feature representation. In summary, the re-weighting operations in the cross-hypothesis max-pooling and the cross-proposal fusion can reduce the negative effects of the inaccurate input proposals.

 With $\bold{f}$ generated by the cross-proposals fusion, we can generate either the category-aware hash representation that consists of $c$ pieces of hash codes, or the semantic hash representation that consists of one piece of hash code. Next we will present these cases separately.
 
 \begin{figure*}
  \centering
  \includegraphics[width=1.0\hsize]{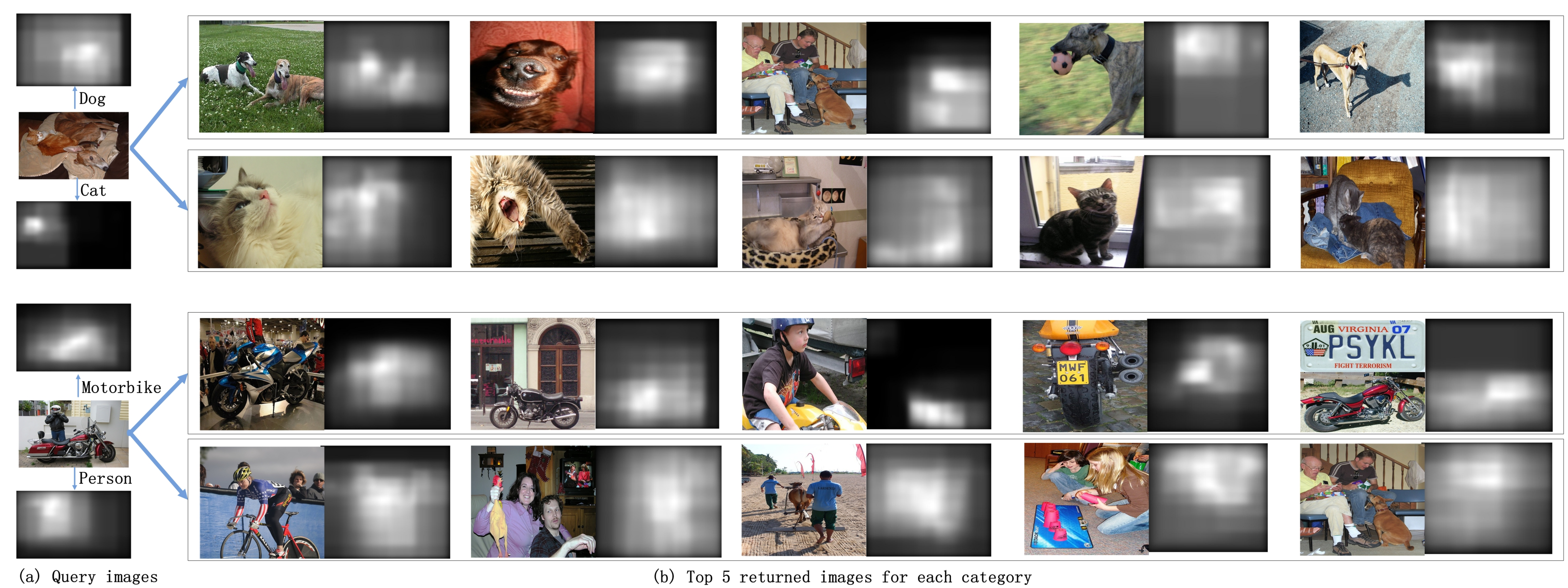}
  \caption{Two examples of the results of category-aware hashing. Each retrieved image has a corresponding grey image of the saliency map, in which whiter color indicates higher possibility of an object existing at that position. Given a query image, the proposed method returns multiple lists of images, each list corresponding to a category. Each image in a returned list is likely to contain a similar object as in the query image, where the approximate location of this object is shown in the corresponding grey image.  }
     \label{image_examples}
\end{figure*}
 
\subsubsection{Category-aware Hash Representation}\label{section_class}
Since $\bold{f}$ is organized in $c$ groups $\bold{f}^{(1)},\bold{f}^{(2)},...,\bold{f}^{(c)}$, each $\bold{f}^{(i)}(i=1,2,...c)$ can be converted into a $b$-bit binary code $\bold{b}^{(i)}=\text{sign}(\bold{f}^{(i)})$, where $\text{sign}(x)=1$ if $x>0$, and otherwise $\text{sign}(x)=0$. For an image $I$, the category-aware hash representation of $I$ is $\bold{b}(I)=(\bold{b}^{(1)}(I),\bold{b}^{(2)}(I),...,\bold{b}^{(c)}(I))$.

To learn this representation, we define $c$ triplet loss functions~\cite{triplet,onestep}, each for one category. 
To obtain triplet samples, we randomly select image $I^+$  and image $I$ that belong to the same category and the negative image $I^-$ is randomly selected from those which do not contain the category. Then we design a triplet loss that tries to preserve the relative similarities in the form: ``image $I$ is more similar to image $I^+$ than to $I^-$''. For the $j$-th category ($j=1,2,...,c$), suppose we have three images $I$, $I^+$ and $I^-$, where both $I$ and $I^+$ belong to the $j$-th category, but $I^-$ does not. Then the triplet loss associated to the $j$-th category is defined by
\begin{equation}
\begin{split}
&\ell_{Triplet}(\bold{f}^{(j)}(I),\bold{f}^{(j)}(I^+),\bold{f}^{(j)}(I^-))\\
&=\max(0,1-||\bold{f}^{(j)}(I)-\bold{f}^{(j)}(I^-)||_2^2 \\
& + ||\bold{f}^{(j)}(I)-\bold{f}^{(j)}(I^+)||_2^2)),\\
\end{split}
\label{local_triplet}
\end{equation}
where $\bold{f}^{(j)}(I)$ represents the vector $\bold{f}^{(j)}$ for the image $I$. 

This loss function is convex. Its sub-gradient with respect to $\bold{f}^{(j)}(I)$, $\bold{f}^{(j)}(I^+)$ and $\bold{f}^{(j)}(I^-)$ can be easily obtain by 
\begin{equation}
\begin{split}
&\frac{\partial \ell_{Triplet}(\bold{f}^{(j)}(I),\bold{f}^{(j)}(I^+),\bold{f}^{(j)}(I^-))}{\bold{f}^{(j)}(I)}\\
& \ \ \ \ \ = 2 ( \bold{f}^{(j)}(I^-) -\bold{f}^{(j)}(I^+) )\\
&\frac{\partial \ell_{Triplet}(\bold{f}^{(j)}(I^+),\bold{f}^{(j)}(I^+),\bold{f}^{(j)}(I^-))}{\bold{f}^{(j)}(I^+)}\\
& \ \ \ \ \ = 2 ( \bold{f}^{(j)}(I^+) -\bold{f}^{(j)}(I) )\\
&\frac{\partial \ell_{Triplet}(\bold{f}^{(j)}(I^-),\bold{f}^{(j)}(I^+),\bold{f}^{(j)}(I^-))}{\bold{f}^{(j)}(I^-)}\\
& \ \ \ \ \ = 2 ( \bold{f}^{(j)}(I) -\bold{f}^{(j)}(I^-) )\\
\end{split}
\end{equation}
when $1-||\bold{f}^{(j)}(I)-\bold{f}^{(j)}(I^-)||_2^2  + ||\bold{f}^{(j)}(I)-\bold{f}^{(j)}(I^+)||_2^2 > 0$. Otherwise the sub-gradients are all zeros.

\subsubsection{Semantic Hash Representation}
For semantic hashing, we assume the target length of a hash code is $q$ bits. We first use a fully-connected layer to convert the $(c\times b)$-dimensional $\bold{f}$ to a $q$-dimensional $\bold{s}$. $\bold{s}$ can be converted to a $q$-bit binary code by $\text{sign}(\bold{s})$, where $\text{sign}(x)=1$ if $x>0$, and otherwise $\text{sign}(x)=0$.

Next we present the triplet loss defined on $\bold{s}$. Since the original triplet loss~\cite{triplet,onestep} is designed for single-label data, here we propose a weighted triplet loss for multi-label data. Specifically, we define the similarity function $sim(I_a,I_b)$ as the number of shared labels between the images $I_a$ and $I_b$. Then, for the images $I$, $I^+$ and $I^-$ and $sim(I,I^+)>sim(I,I^-)$, the weighted triplet loss is defined by
\begin{equation}\label{W-Tri}
\begin{split}
&\ell_{W-Triplet}(\bold{s}(I),\bold{s}(I^+),\bold{s}(I^-))\\
&=(2^{sim(I,I^+)}-2^{sim(I,I^-)})\ell_{Triplet}(\bold{s}(I),\bold{s}(I^+),\bold{s}(I^-))\\
\end{split}
\end{equation}
where $\ell_{Triplet}$ is defined in (\ref{local_triplet}), and $\bold{s}(I)$ is the $q$-dimensional vector for the image $I$. 

In many existing supervised hashing methods, the side information is in the form of pairwise labels indicating the similarites/dissimilarites on image pairs. In these hashing methods, a straightforward way is to define the pairwise loss functions which preserve the
pairwise similarities of images. Some
recent papers (e.g.,~\cite{triplet,onestep}) learn hash functions by using
triplet loss functions, which seek to preserve the relative similarities in the form: ``image $A$ is more
similar to image $B$ than to image $C$''. Such a form of
triplet-based relative similarities can be more easily obtained than
pairwise similarities (e.g., users' click-through data from image
retrieval applications). 

\section{Category-Aware Retrieval\label{IAR}}
Suppose that we have a set of images $\mathcal{S}=\{I_1,I_2,...,I_{|\mathcal{S}|}\}$ with $|\mathcal{S}|$ being the number of images in $\mathcal{S}$ for retrieval. Independently from other categories, for the $j$-th category ($j=1,2,...,c$), we can generate the binary codes $\bold{b}^{(j)}(I_1),\bold{b}^{(j)}(I_2),...,\bold{b}^{(j)}(I_{|\mathcal{S}|})$, and then conduct retrieval based on these codes. Hence, for a query image, the retrieved results can be organized in $c$ groups, where the $j$-th group has a list of images, each of which is likely to contain a similar object of the $j$-th category.

An issue which needs to be considered here is that the number of objects in an image may be less than $c$, and it is inappropriate to organize the retrieved results in $c$ groups for all of the query images. Since the label probability calculation module (see Section \ref{MLCM}) outputs a probability vector $\bold{p}$, where $\bold{p}_j$ represents the predicted value for the possibility of the input image containing objects in the $j$-th category. For the $c$ groups of retrieved results, we can remove the $j$-th group if and only if $\bold{p}_j$ is less than some threshold. In our experiments, we empirically set this threshold to be $0.2$.

For those images in the database for retrieval, each image is first encoded into $c$ pieces of $b$-bit binary codes. Then we collect those hash codes with a probability score (i.e., $\bold{p}_j$ for the $j$-th piece of hash code) no less than $0.2$. We organize the collected hash codes in $c$ groups. The $j$-th group contains the hash codes with each being the $j$-th piece of code of some image. Finally, we build a hash table to store the hash codes in each group, respectively. In retrieval, for a test query image, we first convert it into $c$ pieces of $b$-bit codes, and then remove those codes with a probability score less than $0.2$. For each of the rest hash codes, we conduct search in the corresponding hash table and obtain a list of retrieved images.


Figure~\ref{image_examples} shows two examples of results from our experiments. For the first example, when retrieving with a query image containing a \textit{cat} and a \textit{dog}, the proposed method returns two lists of retrieved images. Each image in the first/second list is likely to have a \textit{cat}/\textit{dog} similar to that in the query image, where the approximate location of this \textit{cat}/\textit{dog} is also indicated (in the associated grey image). The grey images are saliency maps that are obtained in favor of the automatically generated region proposals (see Section \ref{RPG}) and the predicted probability scores (i.e., the vector $\bold{P}$ in Section \ref{MLCM}).

\section{Experiments}
In this section, we evaluate the performance of the proposed method for either semantic hashing or category-aware hashing, and compare it with several state-of-the-art hashing methods.

\subsection{Datasets and Evaluation Metrics}
We evaluate the proposed method on three public datasets of multi-label images: VOC 2007~\cite{everingham2010pascal}, VOC 2012~\cite{everingham2010pascal} and MIRFLICKR-25K~\cite{huiskes2008mir}. \begin{itemize}
\item VOC 2007 consists of 9,963 multi-label images which are collected from Flickr\footnote{http://www.flickr.com/}. There are 20 object classes in this dataset. On average, each image is annotated with 1.5 labels.
\item VOC 2012 consists of 22,531 multi-label images in 20 classes. Since the ground truth labels of the test images are not available, in our experiments, we only use 11,540 images from its training and validation set.

\item MIRFLICKR-25K consists of 25,000 multi-label images downloaded from Flickr. There are 38 classes in this dataset. Each image has 4.7 labels on average.
\end{itemize}

In each dataset, we randomly select 2,000 images as the test query set, and the rest images are used as training samples.  Note that, we only use 11,540 images in VOC2012 dataset, which have the ground truth labels. The number of training samples and testing samples are shown in Table \ref{number_training}:

\begin{table}[h]
\small
    \centering \caption{The number of training samples and testing samples.}
    \begin{tabular}{|c|c|c|c|}
         \hline
         & VOC2007 & VOC2012 & MIRFLICKR-25K \\
        \hline
       $\#$Test & 2,000  & 2,000  & 2,000\\
         \hline 
        $\#$Train & 7,963  & 9,540  & 23,000\\
         \hline
        \end{tabular}
        \label{number_training}
\end{table}

To evaluate the performance, we use four evaluation metrics: Normalized Discounted Cumulative Gains (NDCG)~\cite{jarvelin2002cumulated}, Mean Average Precision (MAP)~\cite{baeza1999modern}, Weighted MAP~\cite{zhao2015deep} and Average Cumulative Gains (ACG)~\cite{jarvelin2000ir}.

NDCG is a popular evaluation metric in information
retrieval. Given a query image, the DCG score at the position $m$ is defined as 
\begin{equation}
\text{DCG}@m = \sum_{j=1}^{m} \frac{2^{r(j)}-1}{\log(1+j)},
\end{equation}
where $r(j)$ is the similarity between the $j$-th position image and the query image, which is defined as the number of shared labels between the query image and the $j$-th retrieved image. Then, the NDCG
score at the position $m$ can be
calculated by NDCG$@m = \frac{\text{DCG}@m}{Z_m}$, where $Z_m$ is the
maximum value of DCG$@m$, making the value of
NDCG fall in the range $[0,1]$.

ACG$@m$ represents the sum of similarities between the query image and each of the top $m$ retrieved images, which can be calculated by
\begin{equation}
\text{ACG}@m = \sum_{j=1}^m \frac{r(j)}{m}.
\end{equation}

MAP is a standard evaluation metric for
information retrieval. It is the mean of averaged precisions
over a set of queries, which can be
calculated by
\begin{equation}
\text{MAP} = \sum_{j=1}^n P@j \times pos(j) / N_{pos},
\end{equation}
where $pos(j)$ is an indicator function. If the image
at the position $j$ is relevant (i.e., it at least shares one label with the query image), $pos(j)$ is 1; otherwise $pos(j)$ is
0. $N_{pos}$ represents the total number of relevant images w.r.t. the query image. $P@j = \frac{N_{pos}(j)}{j}$, where $N_{pos}(j)$ represents the number of relevant images within the top $j$ images.

The weighted MAP is defined as
\begin{equation}
\text{Weighted MAP} = \sum_{j=1}^M \text{ACG}@j \times pos(j) / N_{pos}.
\end{equation}

\begin{table*}[ht!]
    \centering \caption{Comparison results of Hamming ranking w.r.t. different numbers of bits on three datasets. }
    \begin{tabular}{|c|c c c c|c c c c|c c c c|}
        \hline
\multirow{2}{*}{ Methods } & \multicolumn{4}{|c}{VOC 2007 } &\multicolumn{4}{|c}{MIRFLICKR25K } & \multicolumn{4}{|c|}{VOC 2012}\\
& 16 bits & 32 bits & 48 bits & 64 bits & 16 bits & 32 bits & 48 bits& 64 bits & 16 bits & 32 bits & 48 bits & 64 bits \\
  \hline
\multicolumn{13}{|c|}{NDCG@1000}\\
        \hline
        Ours & \bf{ 0.7963} &\bf{  0.8696} & \bf{ 0.8865 }&\bf{  0.8929} &\bf{  0.4725} & \bf{ 0.5245} &\bf{  0.5400} &\bf{  0.5552} &\bf{  0.7939} & \bf{ 0.8385 }&\bf{  0.8566} & \bf{ 0.8573}  \\
         \hline
        One-Stage & 0.7808 & 0.8296 & 0.8446 & 0.8530 & 0.4413 & 0.5096 & 0.5392 & 0.5550 & 0.7540 & 0.8012 & 0.8170 & 0.8224 \\
         \hline
          ITQ-CCA & 0.7704 & 0.8007 & 0.8139 & 0.8146 & 0.4498 & 0.4719 & 0.4866 & 0.4921 & 0.7471 & 0.7759 & 0.7815 & 0.7891\\
          \hline
         ITQ & 0.6848 & 0.6768 & 0.6783 & 0.6766 & 0.3734 & 0.3934 & 0.3966 & 0.3982 & 0.6433 & 0.6371 & 0.6338 & 0.6337\\
         \hline
         SH & 0.5404 & 0.5013 & 0.4796 & 0.4697 & 0.3096 & 0.3046 & 0.2998 & 0.2959 & 0.5157 & 0.4718 & 0.4409 & 0.4238 \\
          \hline
\multicolumn{13}{|c|}{ACG@1000}\\
   \hline
        Ours & \bf{ 0.7065} &\bf{  0.7590} & \bf{ 0.7674} & \bf{ 0.7731} & \bf{ 2.6751} &\bf{  2.8353} & 2.8951 & 2.9282 & \bf{  0.7523} &\bf{  0.7841} &\bf{  0.7972} &\bf{  0.7963}\\
         \hline
       One-Stage & 0.7007 & 0.7323 & 0.7407 & 0.7483 & 2.6083 & 2.8302 & \bf{ 2.9141} & \bf{ 2.9642 } & 0.7141 & 0.7522 & 0.7654 & 0.7705 \\
         \hline
          ITQ-CCA & 0.6418 & 0.6658 & 0.6780 & 0.6779 & 2.4785 & 2.5314 & 2.5964 & 2.6149 & 0.6733 & 0.6971 & 0.7031 & 0.7107\\
          \hline
         ITQ & 0.5823 & 0.5695 & 0.5676 & 0.5661 & 2.1964 & 2.2568 & 2.2650 & 2.2747 & 0.5794 & 0.5715 & 0.5669 & 0.5656  \\
         \hline
         SH & 0.4570 & 0.4218 & 0.4044 & 0.3982 & 1.8394 & 1.7668 & 1.7215 & 1.6894 & 0.4660 & 0.4204 & 0.3947 & 0.3785\\
         \hline
         \multicolumn{13}{|c|}{MAP}\\
          \hline
        Ours &\bf{  0.7997} &\bf{  0.8618}  &\bf{  0.8784} &\bf{  0.8830} & \bf{ 0.7994} & \bf{ 0.8317}& \bf{ 0.8366} & \bf{ 0.8361} & \bf{ 0.7942} & \bf{ 0.8437} &\bf{  0.8617} &\bf{  0.8642 }\\
         \hline
       One-Stage & 0.7488 & 0.7995 & 0.8171 & 0.8259 & 0.7727 & 0.8059  & 0.8136  & 0.8179 & 0.7343 & 0.7870 & 0.8055 & 0.8109\\
         \hline
          ITQ-CCA & 0.6913 & 0.7264 & 0.7404 & 0.7396 & 0.7015 & 0.7053 & 0.7174 & 0.7254 &   0.6952 & 0.7254 & 0.7362 & 0.7427\\
         \hline
         ITQ & 0.5845 & 0.5747 & 0.5769 & 0.5741 & 0.6804 & 0.6822 & 0.6796 & 0.6795 & 0.5715 & 0.5984 & 0.5554 & 0.5549  \\
         \hline
         SH & 0.4432 & 0.4071 & 0.3875 & 0.3799 & 0.6174 & 0.6057 & 0.5994 & 0.5952 & 0.4378 & 0.4184 & 0.3641 & 0.3485\\
         \hline
          \multicolumn{13}{|c|}{Weighted MAP}\\
            \hline
        Ours & \bf{ 0.8566} & \bf{ 0.9255} & \bf{ 0.9449} &\bf{  0.9505 }& \bf{ 2.0877} & \bf{ 2.1926} &\bf{  2.2271} & \bf{ 2.2294} &\bf{  0.8429} &\bf{  0.9005} &\bf{  0.9205} &\bf{  0.9229}\\
         \hline
       One-Stage & 0.8007 & 0.8595 & 0.8794 & 0.8903 & 2.0411 & 2.1584 & 2.1958 & 2.2187 &  0.7798 & 0.8414 & 0.8631 & 0.8698 \\
         \hline
          ITQ-CCA & 0.7325 & 0.7725 & 0.7879 & 0.7866 & 1.7359  & 1.7518 & 1.7982 & 1.8185 & 0.7312 & 0.7666 & 0.7779 & 0.7854 \\
         \hline
         ITQ & 0.6214 & 0.6129 & 0.6163 & 0.6132 & 1.6269 & 1.6403 & 1.6344 & 1.6369 & 0.6051 & 0.5715 & 0.5911 & 0.5906 \\
         \hline
         SH & 0.4723 & 0.4342 & 0.4137 & 0.4051 & 1.4077 & 1.3598 & 1.3324 & 1.3150 & 0.4637 & 0.4205 & 0.3865 & 0.3697 \\
         \hline
        \end{tabular}
    \label{sematic_results}
\end{table*}

\begin{figure*}[ht!]
  \begin{flushleft}
  \centering
  \subfigure[VOC 2007]{\label{NUS-WIDE-a}
   \raisebox{-0.01cm}{\includegraphics[width=0.278\textwidth]{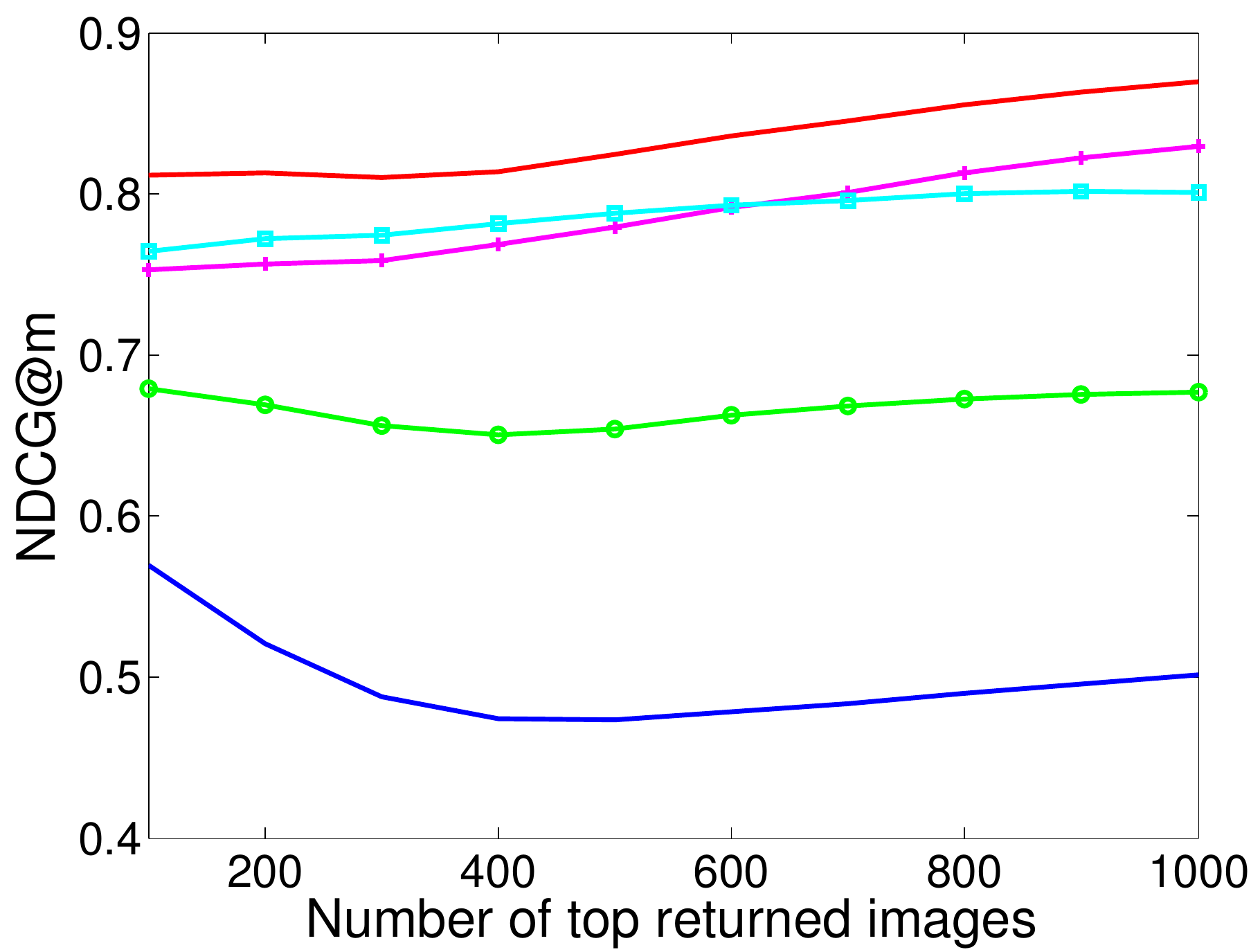}
  }}
  \subfigure[MIRFLICKR]{\label{NUS-WIDE-b}
  \raisebox{-0.01cm}{\includegraphics[width=0.282\textwidth]{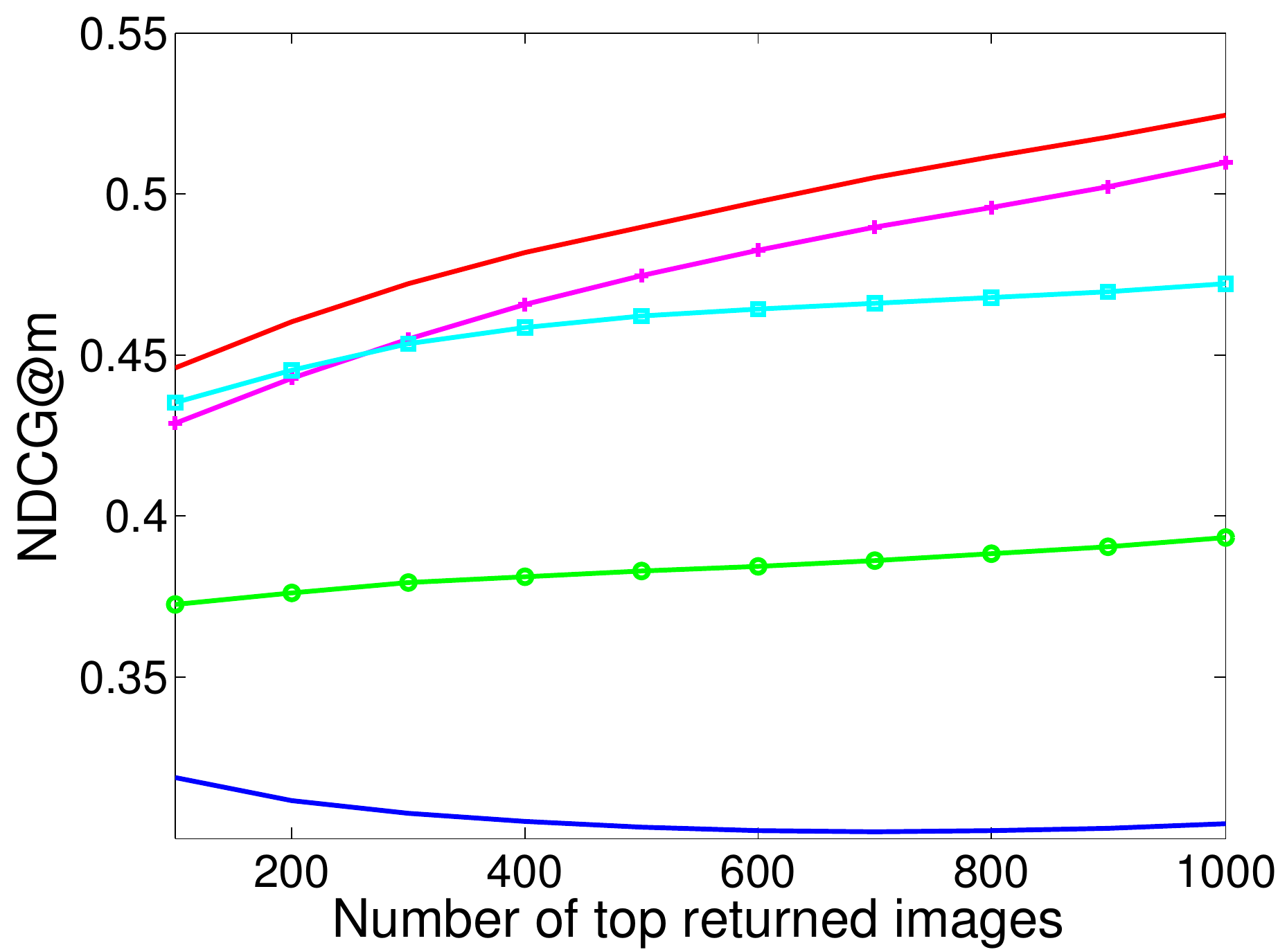}
  }}
  \subfigure[VOC 2012]{\label{NUS-WIDE-c}
  \raisebox{-0.01cm}{\includegraphics[width=0.337\textwidth]{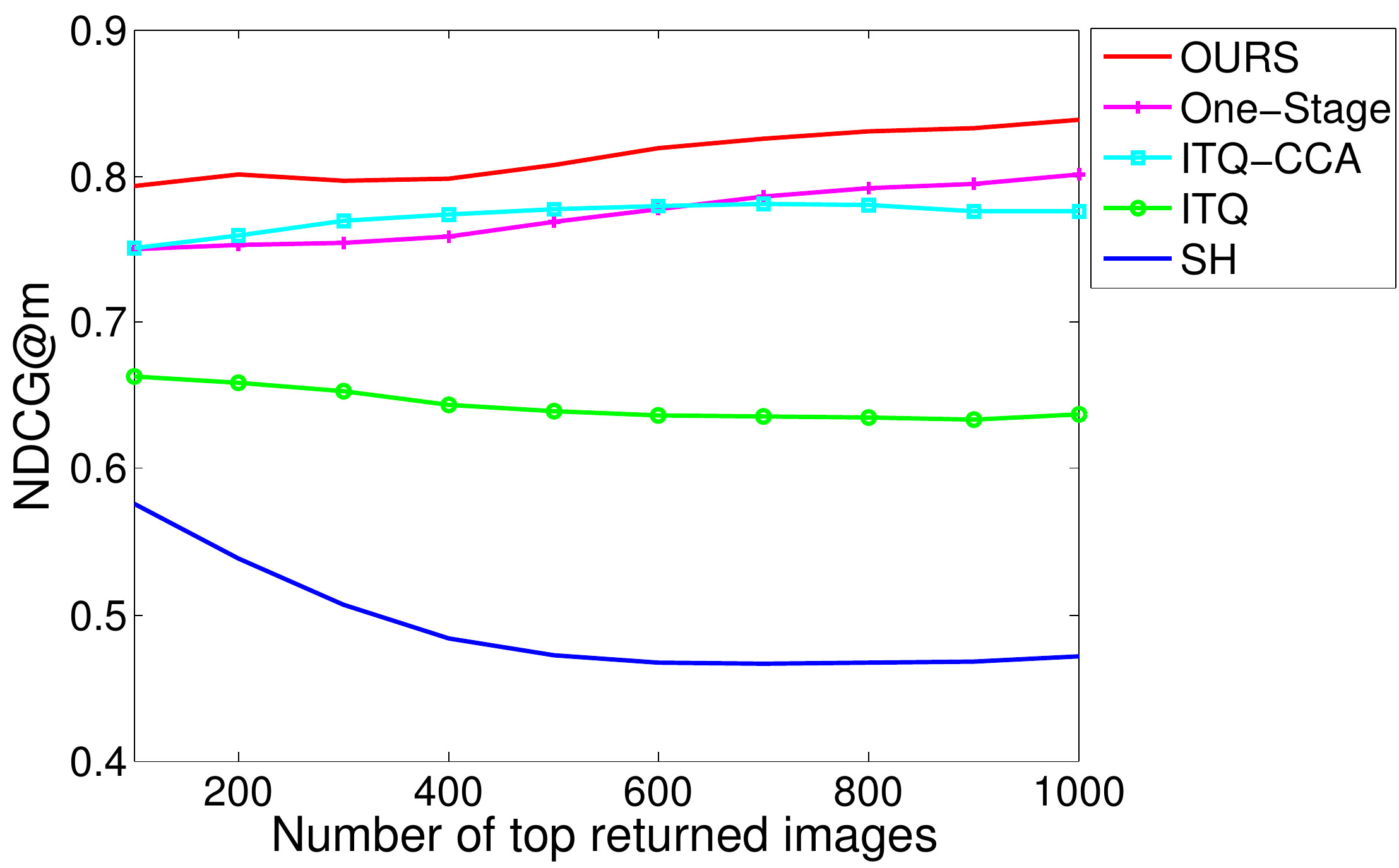}
  }}
  \caption{\footnotesize NDCG curves with 32 bits w.r.t. different numbers of top returned samples.}
  \label{fig: NDCG-result}
  \end{flushleft}
\end{figure*}

\begin{figure*}[ht!]
  \begin{flushleft}
  \centering
  \subfigure[VOC 2007]{\label{NUS-WIDE-a}
   \raisebox{-0.01cm}{\includegraphics[width=0.278\textwidth]{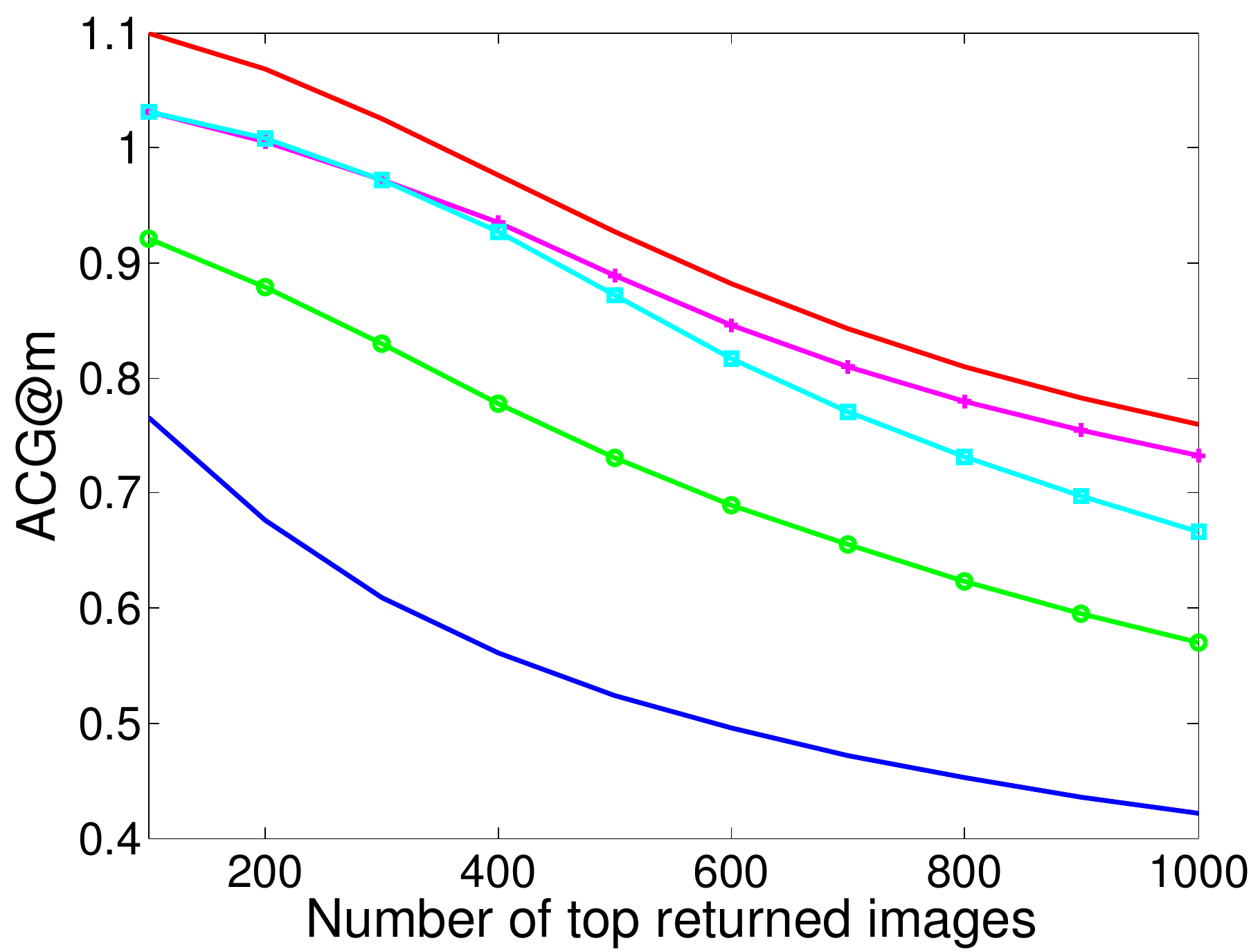}
  }}
  \subfigure[MIRFLICKR]{\label{NUS-WIDE-b}
  \raisebox{-0.01cm}{\includegraphics[width=0.282\textwidth]{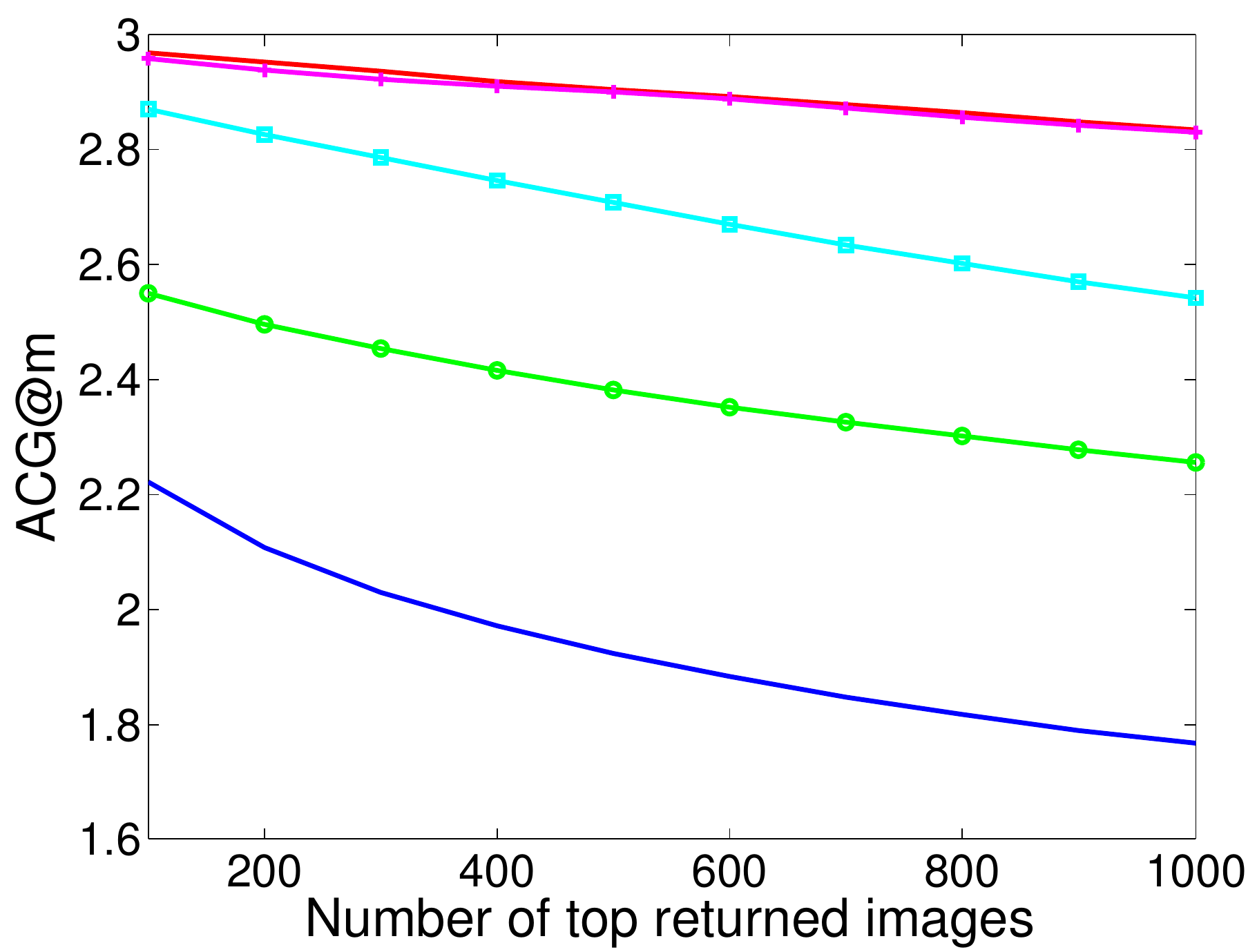}
  }}
  \subfigure[VOC 2012]{\label{NUS-WIDE-c}
  \raisebox{-0.01cm}{\includegraphics[width=0.342\textwidth]{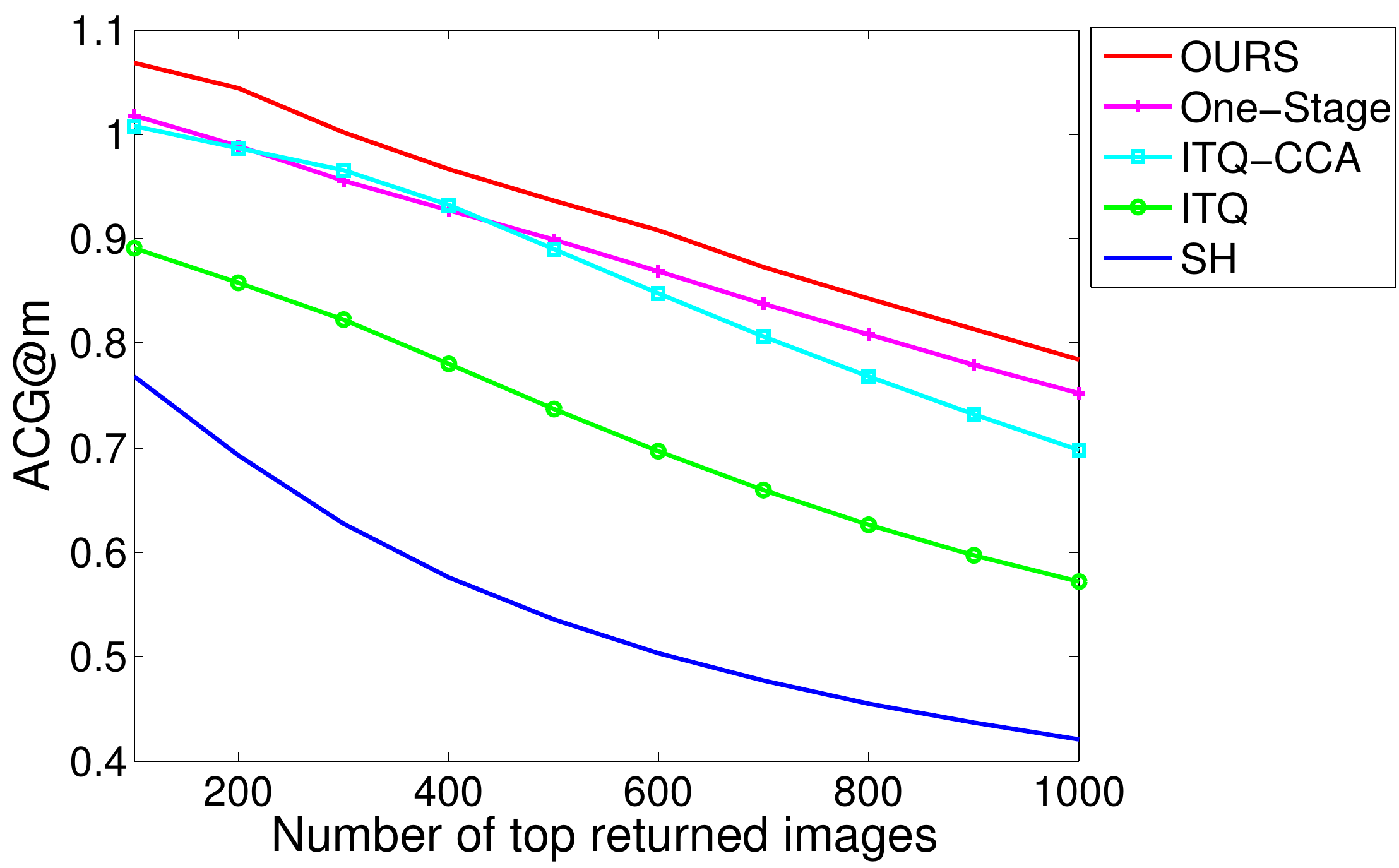}
  }}
  \caption{\footnotesize ACG curves with 32 bits w.r.t. different numbers of top returned samples.}
  \label{fig: ACG-result}
  \end{flushleft}
\end{figure*}

\subsection{Experimental Setting}
We implement the proposed method based on the open-source \textit{Caffe}~\cite{jia2013caffe} framework. The networks are trained using stochastic gradient descent. In training, the weights of the layers are initialized by the pre-trained GoogLeNet model. The base learning rate is set to be 0.0001. After every 30 epochs on the training data, the learning rate is adjusted to one tenth of the current learning rate.
In all of our experiments, we first use GOP to obtain the bounding boxes of region proposals (no more than 1000 proposals for an image). With these bounding boxes, we use non-maximum suppression to obtain a smaller number of boxes, and then select the top $N$ (here we set $N=100$) boxes with the highest confidence. We use the $4$-level pyramid pooling ($4 \times 4, 3 \times 3, 2 \times 2, 1 \times 1)$. The number of feature maps in the last convolution layer is 32, hence the dimension $d$ of each intermediate feature vector is 960 (i.e., $32 \times (4 \times 4 + 3 \times 3 + 2 \times 2 + 1 \times 1)$). The number $b$ of a proposal's feature vectors is set to be the desired hash bits for each category.

\begin{figure*}[t!]
  \centering
  \includegraphics[width=0.9\hsize]{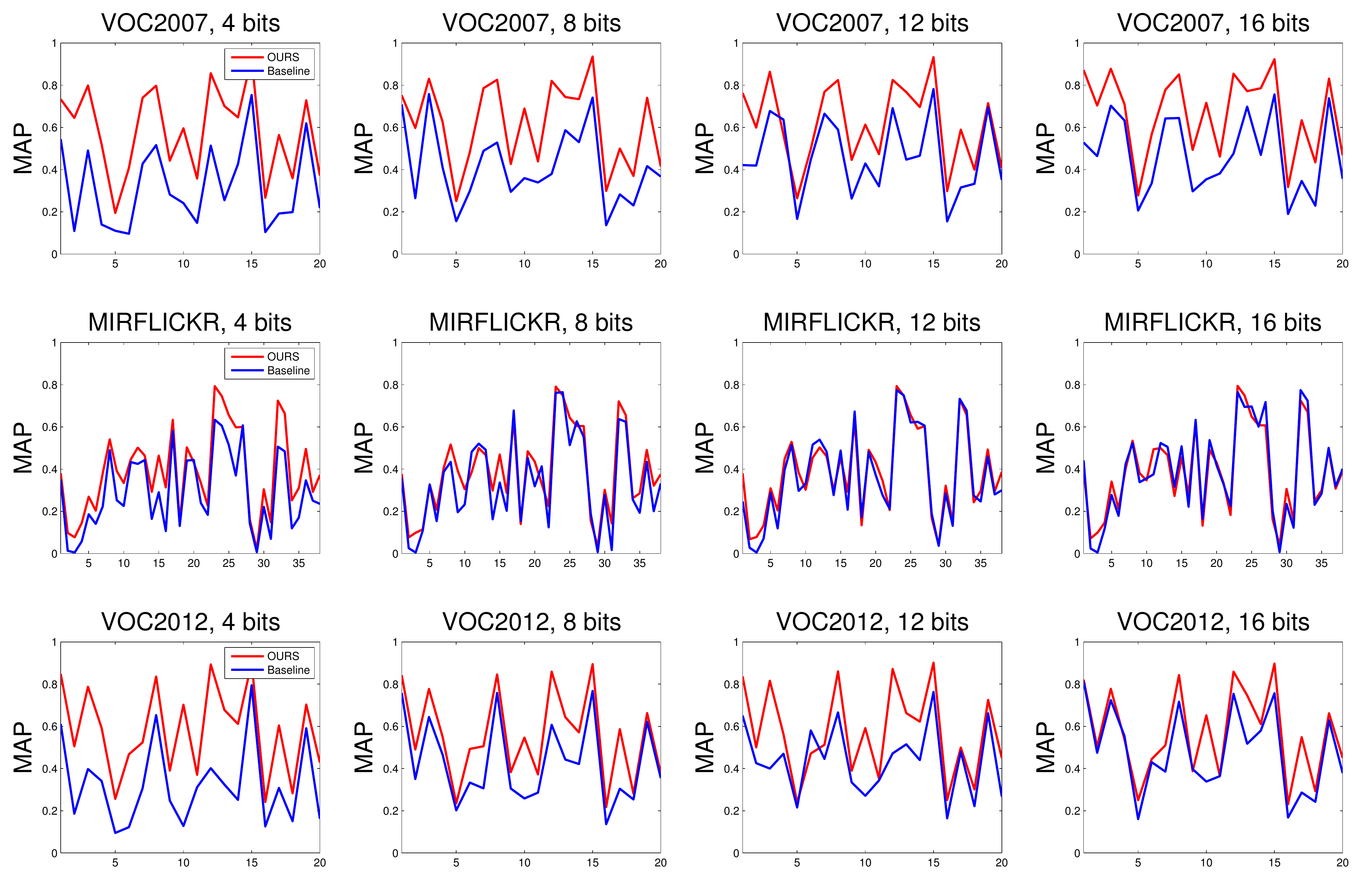}
  \caption{Per-category MAP curves on three datasets. The x-axis represents the categories.}
    \label{all}
\end{figure*}

During training, we use a randomly sampling strategy to generate triplets (i.e., a triplet $(I,I^{+},I^{-})$ of images that $I$ is more similar to $I^{+}$ than to $I^{-}$). Specifically, the proposed network is trained by stochastic gradient descent, where the number of iterations is 15,000 and the mini-batch size is 32. We denote \textit{SharedLabels}$(I_1,I_2)$~\footnote{As an illustrative example, suppose $I_1$ has the class labels $a,b,$ and $c$, $I_2$ has the class labels $a,c,$ and $d$, then we have \textit{SharedLabels}$(I_1,I_2) = 2$ because $I_1$ and $I_2$ have shared labels $a$ and $c$.} as the number of shared labels between the image $I_1$ and the image $I_2$. The procedure of generating triplets in each iteration is shown in the following:
\begin{table}[h]
    \begin{tabular}{l} 
        \textbf{Input}: a batch $S$ of 32 training images. \\
        \textbf{Output}: a set $T$ of triplets. \\
        $T\leftarrow \emptyset$.\\
        \textbf{For} every triplet $t=(I_1,I_2,I_3)$  that $I_1\in S, I_2\in S$ and $I_3  \in S$ \\
        \ \ \ \ \textbf{If} \textit{SharedLabels}$(I_1,I_2)$ $>$ \textit{SharedLabels}$(I_1,I_3)$ \\
        \ \ \ \ \ \ \ \ $T\leftarrow T \bigcup t$\\
         \ \ \ \ \textbf{End If} \\
   \textbf{End For}\\
        \textbf{Output} $T$.\\       
    \end{tabular}
    \label{gen_triplets}
\end{table}

%

The source code of the proposed method is made publicly available at \url{http://ss.sysu.edu.cn/\~py/tip-hashing.rar}.

\begin{figure*}
  \centering
  \includegraphics[width=0.6\hsize]{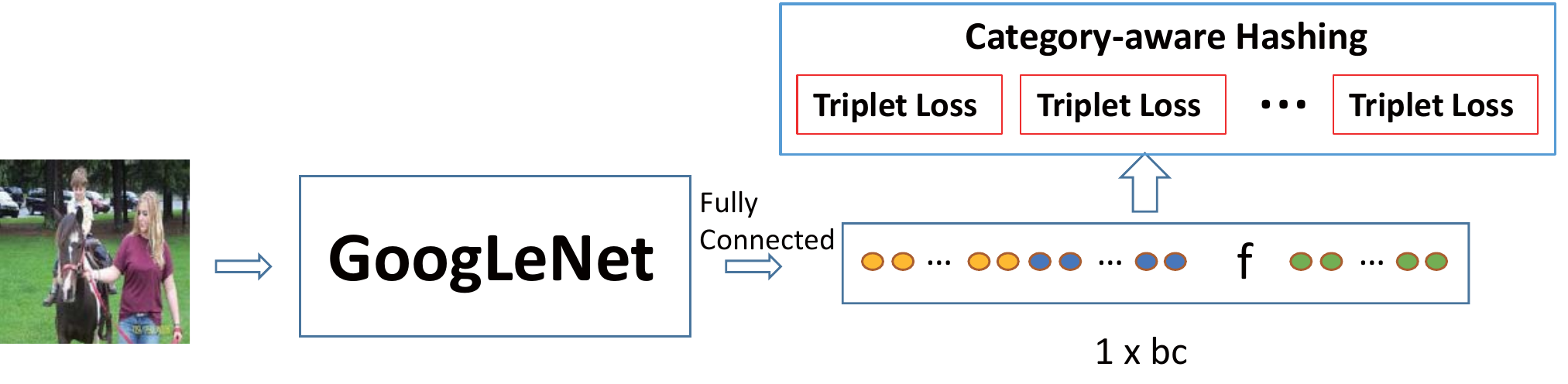}
  \caption{The architecture of the baseline for category-aware hashing.
  }
    \label{baseline}
\end{figure*}

\subsection{Results on Semantic Hashing}
The first set of experiments is to evaluate the performance of the proposed method in semantic hashing.

We use SH~\cite{SH}, ITO~\cite{ITQ}, ITQ-CCA~\cite{ITQ} and One-Stage Hashing~\cite{onestep} as the baselines in our experiments. SH and ITQ are unsupervised methods, while ITQ-CCA and one-stage hashing are supervised methods. One-Stage hashing is a recently proposed deep-networks-based hashing method that is the most related competitor to the proposed method. For SH, ITQ and ITQ-CCA, we use the pre-trained GoogLeNet model\footnote{https://github.com/BVLC/caffe/tree/master/models/bvlc\_googlenet} to extract features for the images. The feature vector for each image is with 1024 dimensions.  For a fair comparison, in our implementation of One-Stage Hashing, we use the architecture of GoogLeNet as its shared sub-network, instead of the NIN architecture used in~\cite{onestep}; we also use the same weighted triplet loss in (\ref{W-Tri}) as the proposed method. The variant of One-Stage Hashing also uses the open-source \textit{Caffe} for training.

Table~\ref{sematic_results} shows the comparison results w.r.t. NDCG@1000, ACG@1000, MAP and Weighted MAP. Figure~\ref{fig: NDCG-result} and Figure~\ref{fig: ACG-result} show the NCCG$@m$ and ACG$@m$ with varying $m$. As can be seen, the proposed method shows superior performance gains over the baselines. On VOC 2007 and VOC 2012, the NDCG@1000 values of the proposed methods
 indicate a $1.9\% \sim 4.9\%$ / $4.2\% \sim 5.2\%$ relative increase over the second best baseline. The ACG@1000 value of the proposed method is 0.7731 with $64$ bits, compared to 0.7483 of One-Stage Hashing. On MIRFLICKR-25K, the values of MAP indicate a relative increase of $2.2 \% \sim 3.4\%$. It can be observed from these results that incorporating automatically generated region proposals and label probability calculation in the process of hash learning can help improve the performance of semantic hashing.

\subsection{Results on Category-aware Hashing}
We also evaluate the performance of the proposed method for category-aware hashing. Since little effort has been devoted to category-aware hashing on multi-label images, to demonstrate the advantages of the proposed method, we implement a deep-networks-based baseline that also outputs $c$ pieces of $b$-bit hash codes, each code corresponding to a category. As shown in Figure \ref{baseline}, this baseline adopts GoogLeNet as the basic framework. After the last (1024-dimensional) fully connected layer of GoogLeNet, a fully connected layer with $c\times b$ nodes is added, and then  this layer is separated into $c$ slices (each is in $b$ dimensions). For the $j$-th ($j=1,2,...,c$) slice,  a triplet loss is defined which regards the images belonging to the $j$-th category as positive examples, and other images as negative ones. To train this baseline, we also use the pre-trained GoogLeNet model to initialize its weights.

The baseline is a simpler category-aware retrieval system, which does not use the region proposal module and label probability module. The experimental results can answer us whether the retrieval system with these two modules can contribute to  the accuracy improvement or not.




For a test query image, we first convert it into $c$ pieces of $b$-bit codes, and then use the hash codes of categories that the test image contains to conduct search in the corresponding hash table and obtain a list of retrieved images.

The MAP results (for each category) are shown in Figure~\ref{all}. We can observe that the proposed method consistently outperforms the baseline. For example, on VOC 2007 with $b = 4$, the averaged MAP (over 20 classes) of the proposed method is 0.5831, compared to 0.3190 of the baseline. On VOC 2012, the averaged MAP of the proposed method has a relative increase of $78.64 \%$ over the baseline with $b = 12$. On MIRFILCKR-25K, the proposed method yields a $12.89 \%$ relative increase over the baseline with $b = 8$ w.r.t. averaged MAP. Figure~\ref{image_examples} shows two examples of results from our experiments.

\section{Conclusions and Future Work}
In this paper, we proposed a deep-networks-based hashing method for multi-label image retrieval, by incorporating automatically generated region proposals and label probability calculation in the hash learning process. In the proposed deep architecture, an input image is converted to an instance-aware representation organized in groups, each group corresponding to a category. Based on this representation, we can easily generate binary hash codes for either semantic hashing or category-aware hashing. Empirical evaluations on both the category-aware hashing and semantic hashing show that the proposed method substantially outperforms the state-of-the-arts.

In future work, we plan to study unsupervised instance-aware image retrieval, in which the virtual classes can be obtained by clustering.
\ifCLASSOPTIONcaptionsoff
  \newpage
\fi



\bibliographystyle{IEEEtran}
\bibliography{paper}
%
%
%

%


\begin{IEEEbiography}[{\includegraphics[width=1in,height=1.25in,clip,keepaspectratio]{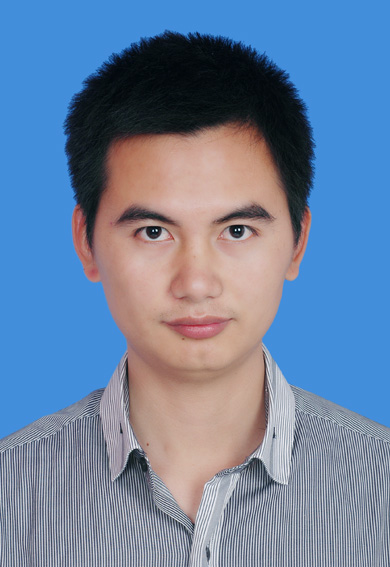}}]{Hanjiang Lai}
	received his B.S. and Ph.D. degrees from Sun Yat-sen University in 2009 and 2014, respectively. He was working as a research fellow at National University of Singapore during 2014-2015. He is now working at Sun Yat-sen university. His research interests includes machine learning algorithms, deep learning, and computer vision.
\end{IEEEbiography}

\begin{IEEEbiography}[{\includegraphics[width=1in,height=1.25in,clip,keepaspectratio]{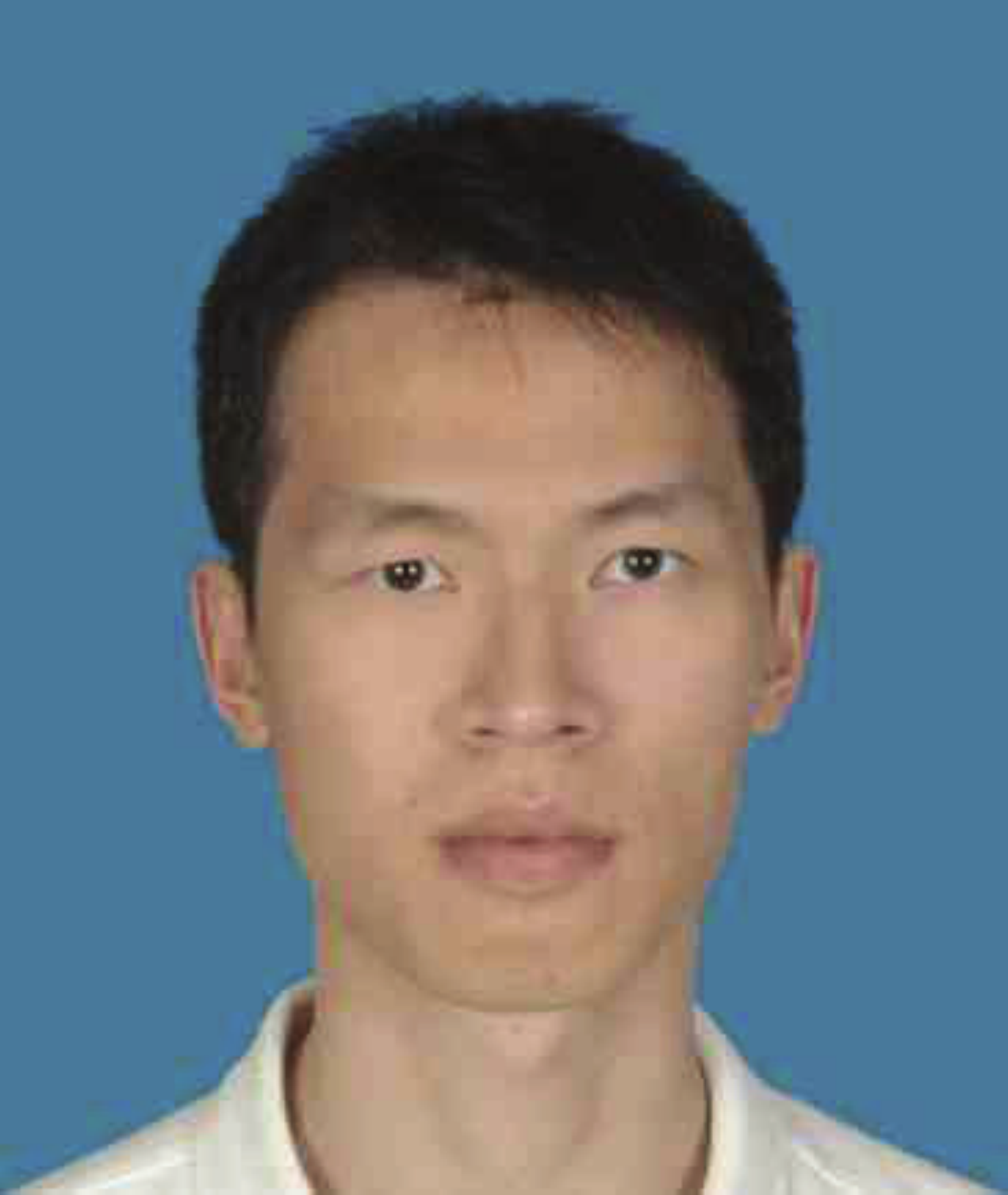}}]{Pan Yan}
	received the B.S. degree in information
science and the Ph.D. degree in computer science
from Sun Yat-sen University, Guangzhou, China,
in 2002 and 2007, respectively.

He is currently an Associate Professor with Sun
Yat-sen University. His current research interests
include machine learning algorithms, learning to
rank, and computer vision.

Dr. Pan has served as a reviewer for several
conferences and journals. He was the winner of the
object categorization task in PASCAL Visual Object
Classes Challenge in 2012.
\end{IEEEbiography}

\begin{IEEEbiography}[{\includegraphics[width=1in,height=1.25in,clip,keepaspectratio]{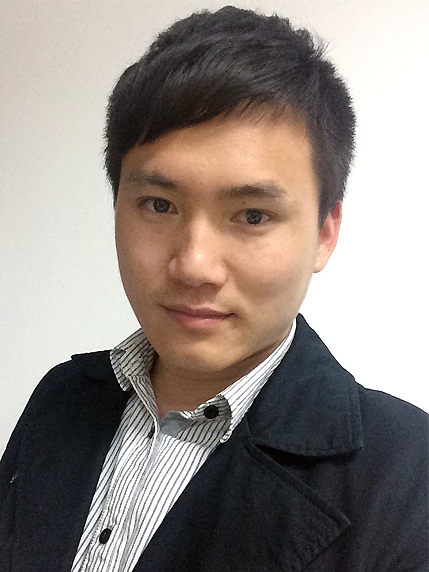}}]{Xiangbo Shu}
	is currently a PhD candidate of School of Computer Science and Engineering, Nanjing University of Science and Technology, Nanjing, China. From Aug. 2014 to present, he is also an visiting scholar in the Department of Electrical and Computer Engineering at National University of Singapore. Her research interests include social multimedia mining, computer vision, and machine learning.
\end{IEEEbiography}

\begin{IEEEbiography}[{\includegraphics[width=1in,height=1.25in,clip,keepaspectratio]{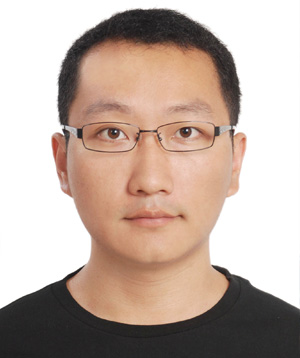}}]{Yunchao Wei}
is a Ph.D. student from the Institute of Information Science, Beijing Jiaotong University, China. He is currently working at National University of Singapore as a Research Intern. His research interests mainly include object classification in computer vision and multi-modal analysis in multimedia.
\end{IEEEbiography}

\begin{IEEEbiography}[{\includegraphics[width=1in,height=1.25in,clip,keepaspectratio]{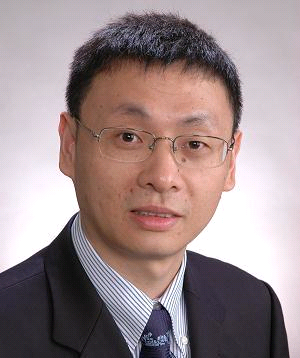}}]{Shuicheng Yan}
		Dr. Yan Shuicheng is currently an Associate Professor at the Department of Electrical and Computer Engineering at National University of Singapore, and the founding lead of the Learning and Vision Research Group (http://www.lv-nus.org). Dr. Yan's research areas include machine learning, computer vision and multimedia, and he has authored/co-authored nearly 400 technical papers over a wide range of research topics, with Google Scholar citation$>$15,000 times. He is ISI highly-cited researcher 2014, and IAPR Fellow 2014. He has been serving as an associate editor of IEEE TKDE, CVIU and TCSVT. He received the Best Paper Awards from ACM MM'13 (Best Paper and Best Student Paper), ACM MM 12 (Best Demo), PCM'11, ACM MM 10, ICME 10 and ICIMCS'09, the runner-up prize of ILSVRC'13, the winner prizes of the classification task in PASCAL VOC 2010-2012, the winner prize of the segmentation task in PASCAL VOC 2012, the honorable mention prize of the detection task in PASCAL VOC'10, 2010 TCSVT Best Associate Editor (BAE) Award, 2010 Young Faculty Research Award, 2011 Singapore Young Scientist Award, and 2012 NUS Young Researcher Award.	
\end{IEEEbiography}




\end{document}